\documentclass{article}

\usepackage{arxiv}

\usepackage[utf8]{inputenc} 
\usepackage[T1]{fontenc}    
\usepackage{hyperref}       
\usepackage{url}            
\usepackage{booktabs}       
\usepackage{amsfonts,amsmath,amssymb}       
\usepackage{nicefrac}       
\usepackage{microtype}      
\usepackage{graphicx}
\usepackage{caption}

\usepackage{color, colortbl}

\usepackage{mathptmx,amsthm}
\usepackage{xcolor}
\usepackage{multirow}
\usepackage{subcaption}
\usepackage{tabularx,adjustbox}

\usepackage{enumitem,makecell}

\newcommand{\etal}{\textit{et al.}}
\newcommand{\ie}{\textit{i.e.,}}
\newcommand{\eg}{\textit{e.g.,}}
\newcommand{\commenttxt}[1]{}

\newcommand{\mybar}{\kern1pt\rule[-\dp\strutbox]{.8pt}{\baselineskip}\kern1pt}

\renewcommand{\mytagline}{} 

\theoremstyle{definition}
\newtheorem{definition}{Definition}[section]

\title{FL-MHSM: Spatially-adaptive Fusion and Ensemble Learning for Flood-Landslide Multi-Hazard Susceptibility Mapping at Regional Scale}
\author{
  Aswathi Mundayatt,
  Jaya~Sreevalsan-Nair\thanks{\texttt{jnair@iiitb.ac.in}} 
  \\
  Graphics-Visualization-Computing Lab,\\
  International Institute of Information Technology Bangalore, Karnataka 560100, India. \\
  \texttt{http://www.iiitb.ac.in/gvcl} \\
}

\begin{document}
\maketitle

\begin{abstract}
Existing multi-hazard susceptibility mapping (MHSM) studies often rely on spatially uniform models, treat hazards independently, and provide limited representation of cross-hazard dependence and uncertainty. To address these limitations, this study proposes a deep learning (DL) workflow for joint flood--landslide multi-hazard susceptibility mapping (FL-MHSM) that combines two-level spatial partitioning, probabilistic Early Fusion (EF), a tree-based Late Fusion (LF) baseline, and a soft-gating Mixture of Experts (MoE) model, with MoE serving as final predictive model. The proposed design preserves spatial heterogeneity through zonal partitions and enables data-parallel large-area prediction using overlapping lattice grids. In Kerala, EF remained competitive with LF, improving flood recall from 0.816 to 0.840 and reducing Brier score from 0.092 to 0.086, while MoE provided strongest performance for flood susceptibility, achieving an AUC--ROC of 0.905, recall of 0.930, and F1-score of 0.722. In Nepal, EF similarly improved flood recall from 0.820 to 0.858 and reduced Brier score from 0.057 to 0.049 relative to LF, while MoE outperformed both EF and LF for landslide susceptibility, achieving an AUC--ROC of 0.914, recall of 0.901, and F1-score of 0.559. GeoDetector analysis of MoE outputs further showed that dominant factors varied more across zones in Kerala, where susceptibility was shaped by different combinations of topographic, land-cover, and drainage-related controls, while Nepal showed a more consistent influence of topographic and glacier-related factors across zones. These findings show that EF and LF provide complementary predictive behavior, and that their spatially adaptive integration through MoE yields robust overall predictive performance for FL-MHSM while supporting interpretable characterization of multi-hazard susceptibility in spatially heterogeneous landscapes.
\end{abstract}


\section{Introduction} \label{sec:intro}
Climate change is reshaping seasonal regimes and intensifying the frequency and magnitude of extremes, thereby increasing the likelihood of compound hazard occurrences. Among these, floods and landslides often interact, as observed in the 2024 Meppadi (Wayanad) flood-landslide event~\cite{vinodini2025mundakkai} and the 2021 Melamchi cascading flood-landslide disaster in Nepal~\cite{sudan_bikash_maharjan_2025_j06m8-z1k87}. Although multi-hazard events represent only 19\% of recorded disasters, they account for 59\% of total economic losses~\cite{united2025global}. These patterns highlight the amplified consequences of compound processes compared with single hazards. In this context, \textit{multi-hazard susceptibility mapping} (MHSM) has emerged as an important risk-assessment tool for estimating the spatial likelihood of interacting hazards using historical occurrences and environmental conditioning factors.

MHSM extends beyond individual hazard assessment by addressing situations in which multiple hazards may co-occur, interact, or intensify one another within the same region in the same timeframe~\cite{gill2014reviewing,piao2022multi}. Existing studies have employed a range of approaches, from GIS-based multi-criteria and statistical methods~\cite{rehman2022multi} to machine learning (ML)~\cite{pourhashemi2025multi}, deep learning (DL)~\cite{ullah2022multi}, hybrid~\cite{bordbar2022multi}, and ensemble models~\cite{kim2025multi}.

Despite this progress, key methodological limitations remain in current MHSM studies. Many existing studies rely on a single predictive model for the entire study region, implicitly assuming that hazard-factor relationships remain uniform across space~\cite{pugliese2024hazard}. However, ecological and physiographic controls often differ across zones, producing region-specific susceptibility behavior and limiting the ability of uniform models to capture localized patterns. In addition, hazards are frequently modeled separately and combined only at a later stage through overlay, weighting, or aggregation~\cite{gill2014reviewing}. Such a concatenation of results is \textit{late fusion}, or \textit{decision-level fusion}. Although such strategies provide a composite view of susceptibility, they do not explicitly represent inter-hazard dependence. Many existing MHSM workflows also remain deterministic, offering limited insight into predictive uncertainty and joint hazard behavior~\cite{tilloy2019review}. Furthermore, in heterogeneous multi-hazard settings, no single modeling strategy may be uniformly effective across all spatial partitions, motivating the need for strategies for the fusion and integration of complementary modeling approaches. In addition, the computational demands of MHSM and the need to manage spatial heterogeneity have led many existing studies to remain highly localized and focused on relatively small study areas. This leaves a gap in extending MHSM to regional scales, where large-area hazard events and stronger spatial heterogeneity must be addressed explicitly.

Recent advances in environmental prediction suggest relevant directions for addressing these limitations. Spatially-adaptive modeling provides a basis for handling spatially varying hazard responses~\cite{li2020spatial}, while joint modeling of correlated outputs enables more explicit representation of inter-hazard dependence and predictive uncertainty~\cite{dai2026multi}. Building on these directions, this study proposes an FL-MHSM  workflow for spatially-adaptive joint multi-hazard modeling and fusion, specifically for floods and landslides, in large-scale regional study areas.

\begin{figure*}[!t]
    \centering
     \includegraphics[width=1\textwidth]{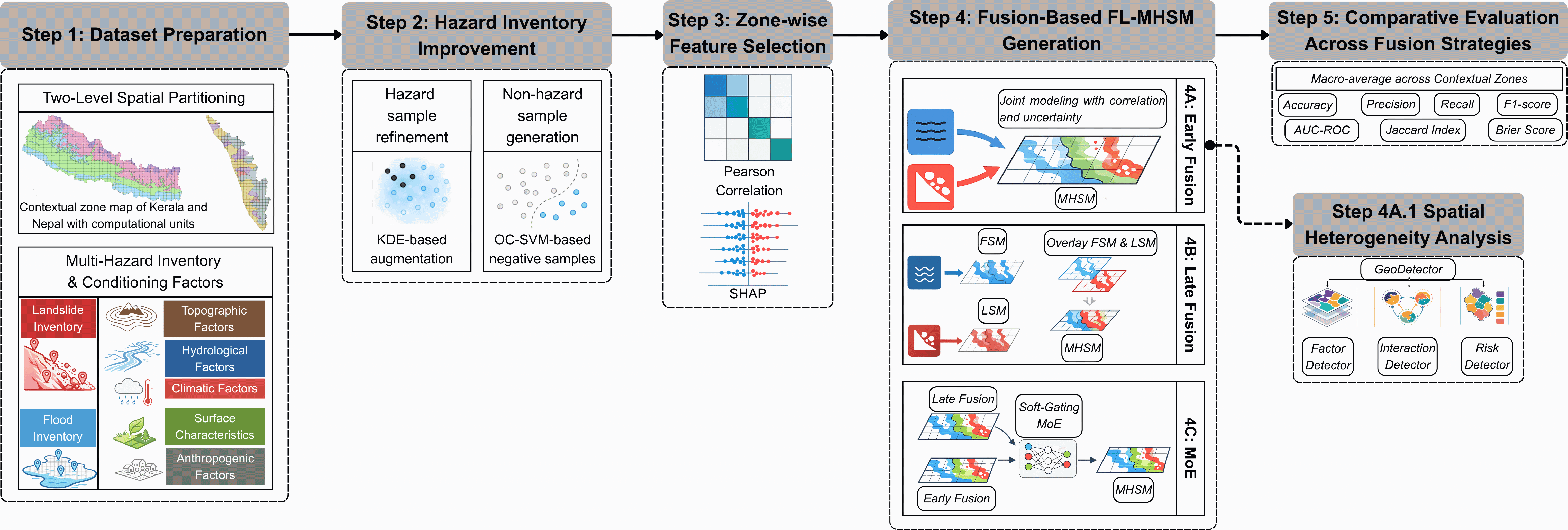}
    \caption{Workflow of the proposed FL-MHSM, consisting of data preparation using contextual zone maps and computational units, EF, LF, and MoE-based susceptibility modeling, and subsequent evaluation and MoE-based spatial heterogeneity analysis.}
    \label{fig:workflow}
\end{figure*}

The main contributions of this study are as follows:
\begin{itemize}[noitemsep, topsep=0pt]
    \item Spatial partitioning strategy based on ecological and physiographic heterogeneity, for \textit{contextual-zone}-specific MHSM across the selected large-scale regional study areas, specifically for a flood-landslide multi-hazard susceptibility mapping (FL-MHSM) workflow.
    \item Probabilistic Early-Fusion (EF) DL model for MHSM using a multivariate Gaussian (MVG) formulation to capture inter-hazard correlation and aleatoric uncertainty.  
    \item Soft-gating Mixture of Experts (MoE) for an ensemble of multiple spatially-adaptive fusion models.
\end{itemize}
The entire workflow of the proposed FL-MHSM is illustrated in Figure~\ref{fig:workflow}.  EF and LF are evaluated as complementary joint and independent modeling strategies, respectively, with LF implemented as an eXtreme Gradient Boosting (XGBoost) baseline. MoE then serves as the final predictive model by integrating these complementary experts in an ensemble, and the MoE-generated MHSMs are further interpreted using GeoDetector to examine spatial heterogeneity. While LF models are conventionally used, we make appropriate modifications to adapt them to our FL-MHSM workflow.

\section{Related Work} \label{sec:relatedwork}
Hazard susceptibility mapping (HSM) has evolved from knowledge-driven and statistical approaches toward more flexible ML and DL workflows. Traditional methods, including expert-based weighting schemes~\cite{swain2020flood,myronidis2016landslide,pourghasemi2012application, salehpour2021gis,akay2021flood,arumugam2023gis} and empirical statistical models~\cite{rahmati2016flood,vakhshoori2016landslide,tang2021comparison,khosravi2016flash,kavzoglu2014landslide,eiras2021discriminant}, are often computationally efficient and interpretable, but are generally less effective in representing complex nonlinear relationships in hazard environments. The increasing availability of high-resolution Earth Observation (EO) data, together with advances in computational capability, has accelerated the use of ML and DL methods in HSM. A wide range of algorithms, including Support Vector Machine (SVM), Random Forest (RF), XGBoost (XGB), Light Gradient Boosting Machine (LightGBM)~\cite{kurugama2024comparative}, Artificial Neural Network (ANN), and Convolutional Neural Network (CNN)~\cite{ado2022landslide, seydi2022comparison}, has been applied for susceptibility estimation, with tree-based ensembles and neural models often showing strong performance in representing nonlinear environmental relationships~\cite{usta2024comparison,shahri2019landslide,selamat2022landslide,khoirunisa2021gis}. Ensemble methods, including Stacking, Bagging, and Boosting~\cite{hu2020landslide,kadavi2018application} and hybrid workflows combining ML or DL models with statistical methods, have demonstrated that combining complementary learners can improve susceptibility prediction beyond single-model implementations~\cite{tehrany2014flood,islam2021flood,shafizadeh2018novel}.

Many methodological advances originally developed for single-hazard susceptibility mapping (SHSM) have increasingly been adapted for MHSM. Existing MHSM studies include statistical-ML hybrid models, such as generalized linear models combined with RF~\cite{pourhashemi2025multi}, generalized additive models for cryospheric multi-hazard analysis~\cite{nicu2023multi}, and ensemble learning workflows based on RF, XGB, and Boosted Regression Trees (BRT) for combinations of drought, flood, wildfire, and related hazards~\cite{piao2022multi,kim2025multi}. Multi-criteria decision analysis (MCDA) approaches, particularly GIS-AHP, have also been used for integrated assessment of hazards such as soil erosion, torrential flooding, landslides, and tsunamis~\cite{krassakis2023multi,rehman2022multi}. More recently, DL models such as CNNs and coupled models combining tree-based and neural models have also been explored for multi-hazard settings~\cite{ullah2022multi,yu2024susceptibility}. These studies demonstrate the growing methodological diversity in MHSM; however, joint probabilistic treatment of correlated or co-occurring hazards remains relatively underexplored in the state of the art.

Current MHSM studies, related to fusion, commonly follow three main directions: (i) to generate individual hazard susceptibility maps separately and then combine them into a final multi-hazard product through weighted integration approaches such as AHP-, FR--AHP-, or GIS-based weighted synthesis~\cite{rehman2022multi}, (ii) to evaluate multiple models separately for each hazard, retain the best-performing hazard-specific model, and then integrate the selected outputs to construct the final MHSM~\cite{kim2025multi}, and (iii) to use a rule-based fusion, which is rare and moves beyond simple weighted overlay, \eg~combining separately produced hazard layers through a Mamdani fuzzy inference system~\cite{karakas2023hybrid}.

Recent studies in susceptibility assessment have explored spatially-adaptive formulations to account for location-dependent variation in the influence of conditioning factors. A geographically weighted regression (GWR) model addressed this issue in landslide susceptibility mapping~\cite{li2020spatial}, showing that local modeling can better represent spatially varying hazard-factor relationships than global formulations. Applications in environmental susceptibility studies include the use of spatially-adaptive geographically weighted regression for capturing localized erosion drivers, which outperforms the conventional stationary models~\cite{peng2025spatially}. Beyond local regression, regional division strategies have also been explored to account for heterogeneity across subareas. Dividing the study area into local regions and evaluating subregional-scale susceptibility yielded results more consistent with observed hazard distributions than a traditional global model~\cite{ma2023evaluation}. Similarly, spatially-adaptive ML has been implemented by first identifying spatially homogeneous clusters and then constructing locally optimal models for each cluster, showing that different subregions may require different predictive models under spatial heterogeneity~\cite{wang2023spatially}. Recently, spatial heterogeneity zoning combined with zone-specific feature optimization has been shown to improve susceptibility prediction and clarify subregional controlling mechanisms~\cite{zhao2025impact}. Extending this idea to ML, spatial factor optimization can be integrated with a geographically weighted random forest model so that local ML accommodates spatial differentiation in landslide conditioning processes~\cite{lu2025geographically}. Collectively, these studies indicate that spatially-adaptive susceptibility assessment has been implemented through both local regression and zone-specific modeling strategies.

In parallel with spatially-adaptive modeling, environmental prediction studies have also explored joint multivariate models for modeling related outputs through shared dependency structure. Multi-output Gaussian process formulations provide a principled probabilistic basis for learning correlated targets through structured covariance functions, allowing information transfer across outputs~\cite{liu2018remarks}. In environmental monitoring, multi-output Gaussian regression has improved predictive accuracy by jointly estimating correlated soil properties rather than treating them independently~\cite{dai2026multi}. Multivariate spatial Bayesian models have similarly been used for simultaneous prediction of multiple air pollutants, showing that cross-variable spatial covariance can improve robustness and uncertainty quantification compared with separate models~\cite{gong2021multivariate}. Related ideas have also been explored in DL, where multivariate generative architectures have been developed to model the joint distribution of correlated environmental variables~\cite{dey2024predicting}. In remote sensing, multi-output regression has further shown that exploiting inter-parameter relationships can yield more stable and accurate estimates than independent single-output prediction~\cite{tuia2011multioutput}. Taken together, these studies provide methodological support for extending joint probabilistic multi-output modeling to MHSM settings involving correlated hazard processes.

\begin{table}[t]
\centering
\caption{Multi-source datasets used in the proposed FL-MHSM workflow.}
\renewcommand{\arraystretch}{1.2}

\begin{tabularx}{\linewidth}{@{}p{3cm} p{4cm} X@{}}
\hline
Category & Raw Data & Source \\
\hline

\multirow{2}{*}{Contextual Zone Map}
 & Kerala zonation map & Madhav Gadgil Report / WGEEP ESZ classification~\cite{gadgil2011report} \\
 & Nepal zonation map & NNH categories associated with the RESOLVE ecoregions dataset~\cite{dinerstein2017ecoregion} \\
\hline

\multirow{3}{*}{Hazard Inventory}
 & Kerala landslide inventory & NRSC--GSI--KSDMA inventory refined by Hao \etal~\cite{hao2020constructing} \\
 & Nepal landslide inventory & COOLR dataset by Juang \etal~\cite{juang2019using} \\
 & Flood inventory (Kerala and Nepal) & Sentinel-1 SAR flood inventory~\cite{unspider_flood_gee} \\
\hline

\multirow{6}{*}{Topographic}
 & Elevation & NASA SRTM Digital Elevation Model (DEM) \\
 & Aspect & Derived from DEM (NASA SRTM) \\
 & Slope & Derived from DEM (NASA SRTM) \\
 & TWI & Derived from flow direction, flow accumulation, and slope (ArcGIS Pro) \\
 & Landform & Derived from DEM (NASA SRTM) \\
 & SPI & Derived from flow accumulation and slope \\
\hline

\multirow{3}{*}{Hydrological}
 & River Density & Derived from HydroSHEDS river network (ArcGIS Pro) \\
 & Distance to River & Derived from HydroSHEDS river network (ArcGIS Pro) \\
 & Distance to Glacier Lakes & Derived from glacier lake inventory map \\
\hline

\multirow{2}{*}{Climatic}
 & Precipitation & CHIRPS precipitation dataset \\
 & Maximum Temperature & TerraClimate dataset \\
\hline

\multirow{2}{*}{Surface Characteristics}
 & NDVI & Derived from Sentinel-2 Bands B4 and B8 \\
 & LULC & ESA WorldCover \\
\hline

Anthropogenic
 & Population Density & GPWv4.11 Population Density dataset \\
\hline

\end{tabularx}

\label{tab:datasets}
\end{table}

\section{Study Area and Dataset} \label{sec:studyarea}
Here, we provide details of large-scale regional study areas used for our implementation of FL-MHSM, and the description of the multi-source datasets (Table~\ref{tab:datasets}) needed in the workflow.

\subsection{Study Area}
Kerala, located in southwestern India between 8$^\circ$17$^{\prime}$N–12$^\circ$47$^{\prime}$N and 74$^\circ$52$^{\prime}$E–77$^\circ$24$^{\prime}$E, spans an area of 38,863 km\textsuperscript{2} across 14 districts. The state features a tropical monsoon climate with an average annual rainfall of $\sim$2,900 mm and temperatures ranging from 25$^\circ$C to 30$^\circ$C. Its physiography includes lowlands, midlands, and the steep Western Ghats, with underlying geology comprising mainly Precambrian crystalline rocks. Floods are prevalent in the low-lying coastal and midland regions, while landslides primarily affect the orographic slopes of the Western Ghats, exacerbated by deforestation and unplanned development. The 2018 and 2019 extreme monsoon events led to widespread flooding and numerous landslides~\cite{vijaykumar2021kerala}, indicating Kerala's multi-hazard vulnerability.

Nepal, a landlocked Himalayan country (26$^\circ$20$^{\prime}$N–30$^\circ$26$^{\prime}$N, 80$^\circ$03$^{\prime}$E–88$^\circ$12$^{\prime}$E), covers 147,516 km\textsuperscript{2} across 77 districts. The terrain spans from the Terai plains ($\sim$60 m elevation) in the south to the Himalayan range (8,848 m) in the north, divided into Terai, Hilly, and Mountain ecological zones. Nepal's subtropical to alpine climate brings 1,500–2,500 mm of annual rainfall, mostly during the June–September monsoon. The region is prone to deep-seated landslides, especially in the central and eastern hill districts, driven by fragile geology (phyllites, schists) and tectonic activity~\cite{maskey2020glacial}. In addition, the Terai is highly flood-prone due to overflowing river systems, while glacier melt has increased risks of glacial lake outburst floods (GLOFs) and high-altitude landslides~\cite{hu2022temporal}.

\subsubsection{Choice of Regions}
Kerala and Nepal are specifically chosen for the variety in their geography, terrain, climatic characteristics, ecology, relief, and location. Thus, implementing the FL-MHSM workflow on two diverse regions demonstrates its generalizability. Another key characteristic is the aspect ratio of the two regions based on their geographical dimensions. Kerala has a width-to-height aspect ratio of 0.2:1, and thus is a vertical layout. On the contrary, Nepal has an aspect ratio between 3.5:1 and 5:1, giving a horizontal layout. This diversity in aspect ratio demonstrates the generalizability of our proposed spatial partitioning strategy.

\subsection{Dataset}
For implementing FL-MHSM, three critical datasets are required for the selected study areas: contextual zone maps, hazard inventory maps, and rasterized conditioning factors.

\subsubsection{Contextual Zone Map}
Each study area is spatially partitioned into contiguous subregions using a contextual zone map to support localized modeling.

\begin{definition}[Contextual Zone Map]
A contextual zone map of a region $\mathcal{R}$ is a spatial partitioning that delineates contiguous subregions or \textit{zones}, in such a way that there is a local contextual homogeneity within each zone, and a global spatial heterogeneity across all zones in $\mathcal{R}$. The context is based on thematic similarities, for instance, in environment, ecology, geological characterization (\eg~seismic zone map), etc.
\end{definition}

\begin{itemize}[noitemsep, topsep=0pt]
\item {Kerala:}  
Here, the contextual zone map of interest is that of the Ecologically Sensitive Zones (ESZs) defined in the 2011 Madhav Gadgil Report~\cite{gadgil2011report}. The Western Ghats Ecology Expert Panel (WGEEP) identified the Western Ghats as an Ecologically Sensitive Area (ESA) and classified it into three ecological-sensitivity levels: ESZ1, ESZ2, and ESZ3. Among these, ESZ1 represents the most ecologically critical regions, ESZ2 indicates moderate ecological sensitivity, and ESZ3 represents comparatively lower ecological sensitivity. Areas outside the ESZ1--ESZ3 classifications are treated as Non-ESZ regions.

\item {Nepal:}  
Here, the contextual zone map of interest is that of the Nature Needs Half (NNH) classification assigned to the RESOLVE ecoregions dataset (2017 update), where each ecoregion is categorized according to the proportion of protected areas and remaining natural habitat~\cite{dinerstein2017ecoregion}. Within the study area, Category 2 (Nature Could Reach Half), Category 3 (Nature Could Recover), and Category 4 (Nature Imperiled) are present, while Category 1 (Half Protected) does not occur. Areas outside these categories are treated as Non-NNH regions.
\end{itemize}

\subsubsection{Multi-Hazard Inventory Map}
The flood inventories for Kerala and Nepal are derived from Sentinel-1 SAR imagery using a multi-temporal change detection approach implemented in Google Earth Engine, following the principles of the UN-SPIDER radar-based flood mapping workflow~\cite{unspider_flood_gee}. Pre- and post-flood image composites of flood events are compared to identify inundated areas, while permanent water bodies and steep terrain were masked to reduce misclassification. The resulting binary flood extents were converted into vector points and used as flood inventory data for susceptibility modeling.

The landslide inventory for Kerala is derived from the publicly available dataset developed by Hao \etal~\cite{hao2020constructing} for the 2018 monsoon-triggered disaster. This dataset is constructed by integrating two complementary sources: (i) object-based image analysis (OBIA) of Resourcesat-2 and Sentinel-2 imagery conducted by the National Remote Sensing Centre (NRSC), and (ii) field-mapped landslides surveyed by the Geological Survey of India (GSI) in collaboration with the Kerala State Disaster Management Authority (KSDMA).

The landslide inventory for Nepal is obtained from the Cooperative Open Online Landslide Repository (COOLR), developed by NASA and introduced by Juang \etal~\cite{juang2019using}. COOLR integrates NASA’s Global Landslide Catalog (GLC) with crowdsourced contributions collected through the Landslide Reporter interface. These records were harmonized within the COOLR schema and manually validated to form the landslide inventory used in this study.

\begin{figure*}[tbp]
    \centering
    \includegraphics[width=0.7\textwidth]{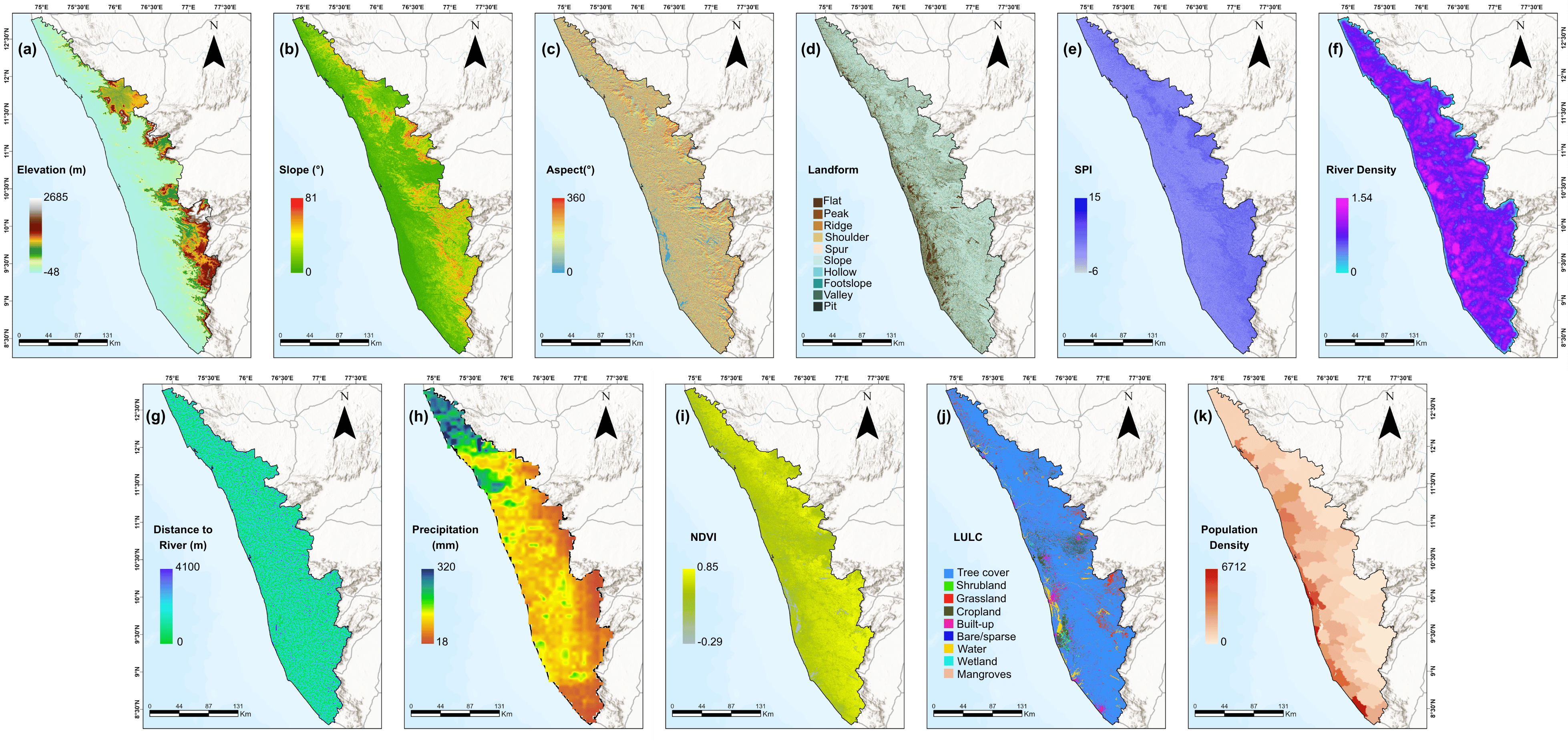}
    \caption{Multi-hazard conditioning factors of the proposed FL-MHSM for Kerala: (a) Elevation, (b) Slope, (c) Aspect, (d) Landform, (e) SPI, (f) River Density, (g) Distance to River, (h) Precipitation, (i) NDVI, (j) LULC, and (k) Population Density.}
    \label{fig:kerala_conditioning_factors}
\end{figure*}

\begin{figure*}[tbp]
    \centering
    \includegraphics[width=0.8\textwidth]{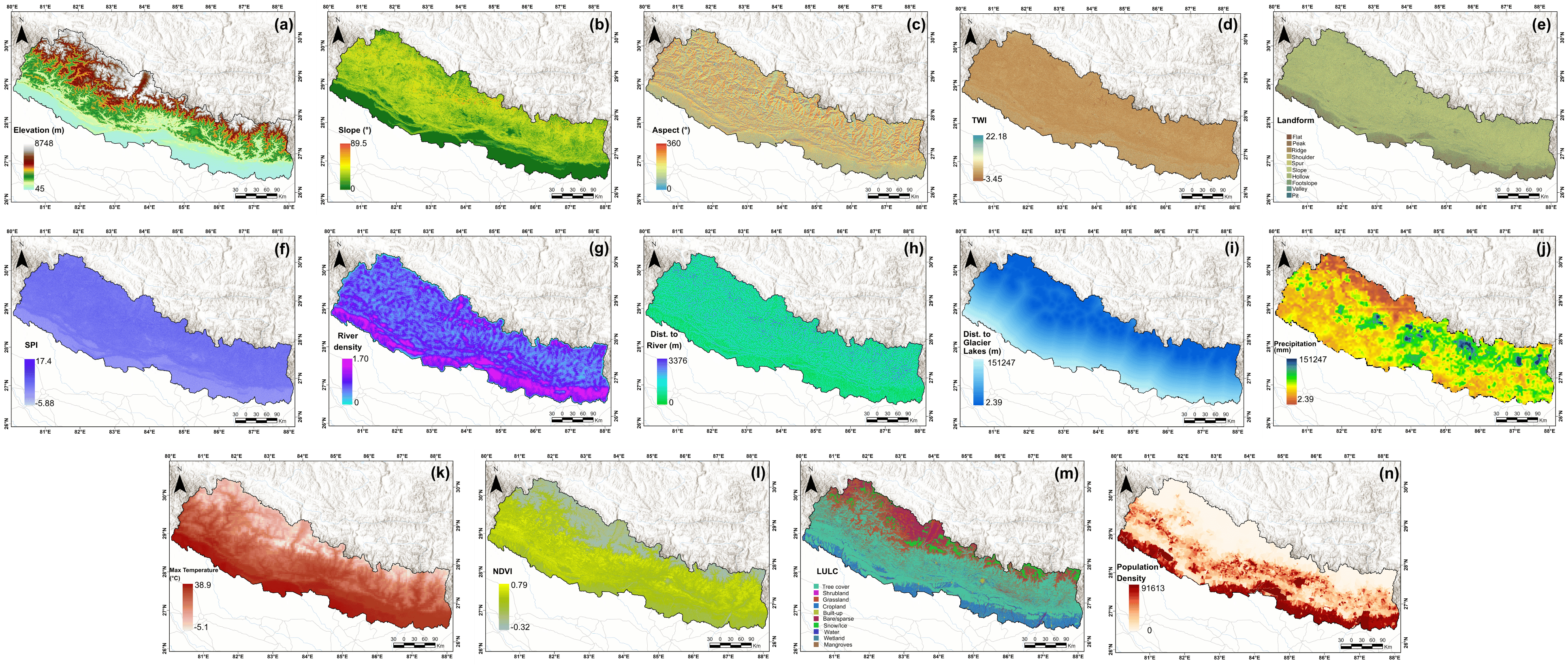}
    \caption{Multi-hazard conditioning factors of the proposed FL-MHSM for Kerala: (a) Elevation, (b) Slope, (c) Aspect, (d) TWI, (e) Landform, (f) SPI, (g) River Density, (h) Distance to River, (i) Distance to Glacier lakes, (j) Precipitation, (k) Max Temperature, (l) NDVI, (m) LULC, and (n) Population Density.}
    \label{fig:nepal_conditioning_factors}
\end{figure*}

\subsubsection{Multi-Hazard Conditioning Factors}
Conditioning factors are selected separately for Kerala and Nepal based on prior studies and regional characteristics. The selected variables represent multi-grouped conditions relevant to susceptibility assessment~\cite{kaya2023parameters}. The conditioning factors are grouped into five thematic categories: topographic, hydrological, climatic, surface characteristics, and anthropogenic. The conditioning factors, with a focus on floods and landslides, are:
\begin{itemize}
\item Topographic variables: Elevation, aspect, slope, topographic wetness index (TWI), stream power index (SPI), and landform.
\item Hydrological factors: River density, distance to river, and distance to glacial lakes.
\item Climatic variables: Precipitation and maximum temperature.
\item Surface characteristics: Normalized difference vegetation index (NDVI) and land use/land cover (LULC).
\item Anthropogenic variables: Population density.
\end{itemize}

Among these variables, TWI, maximum temperature, and distance to glacial lakes were used only for Nepal to account for hydrological and temperature-related processes specific to the Himalayan terrain. The remaining factors were used consistently across both Kerala and Nepal. Figures~\ref{fig:kerala_conditioning_factors} and~\ref{fig:nepal_conditioning_factors} show the conditioning factors used for Kerala and Nepal, respectively.

\section{Methodology} \label{sec:methodology}
This section presents the stages of the proposed FL-MHSM workflow, \ie~the spatial partitioning strategy, contextual-zone-specific feature selection, the EF and LF modeling approaches, the soft-gating MoE formulation, and the analysis procedures used for interpretation and evaluation. The overall workflow of this study is shown in Figure~\ref{fig:workflow}, and schematic of the models is in Figure~\ref{fig:spatialpartition}. The workflow is implemented specifically for flood--landslide multi-hazard susceptibility mapping in the selected study areas.

\subsection{Two-Level Spatial Partitioning Strategy} \label{sec:spatialpart}
\begin{figure*}[!t]
    \centering
     \includegraphics[width=0.4\textwidth]{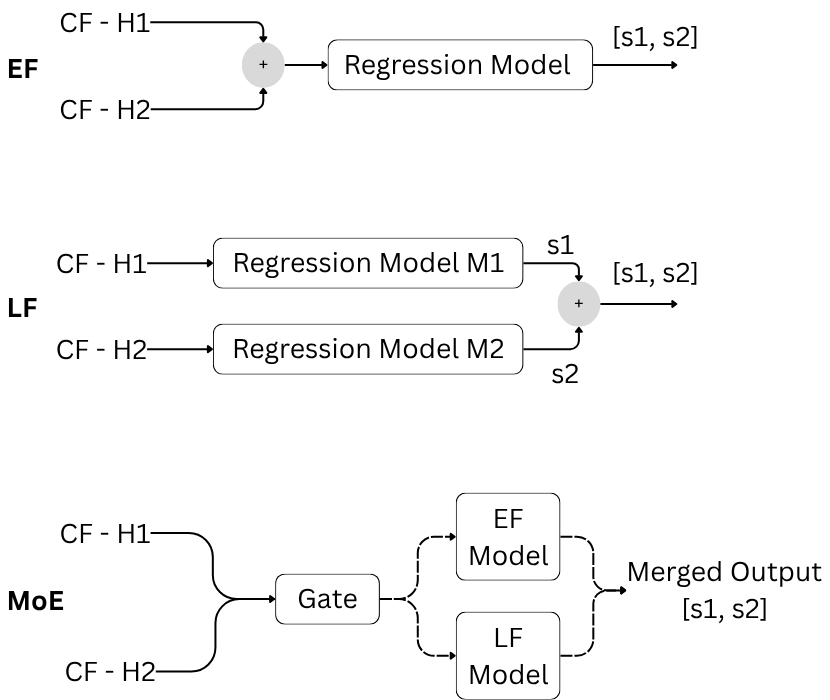}
     \includegraphics[width=0.55\textwidth]{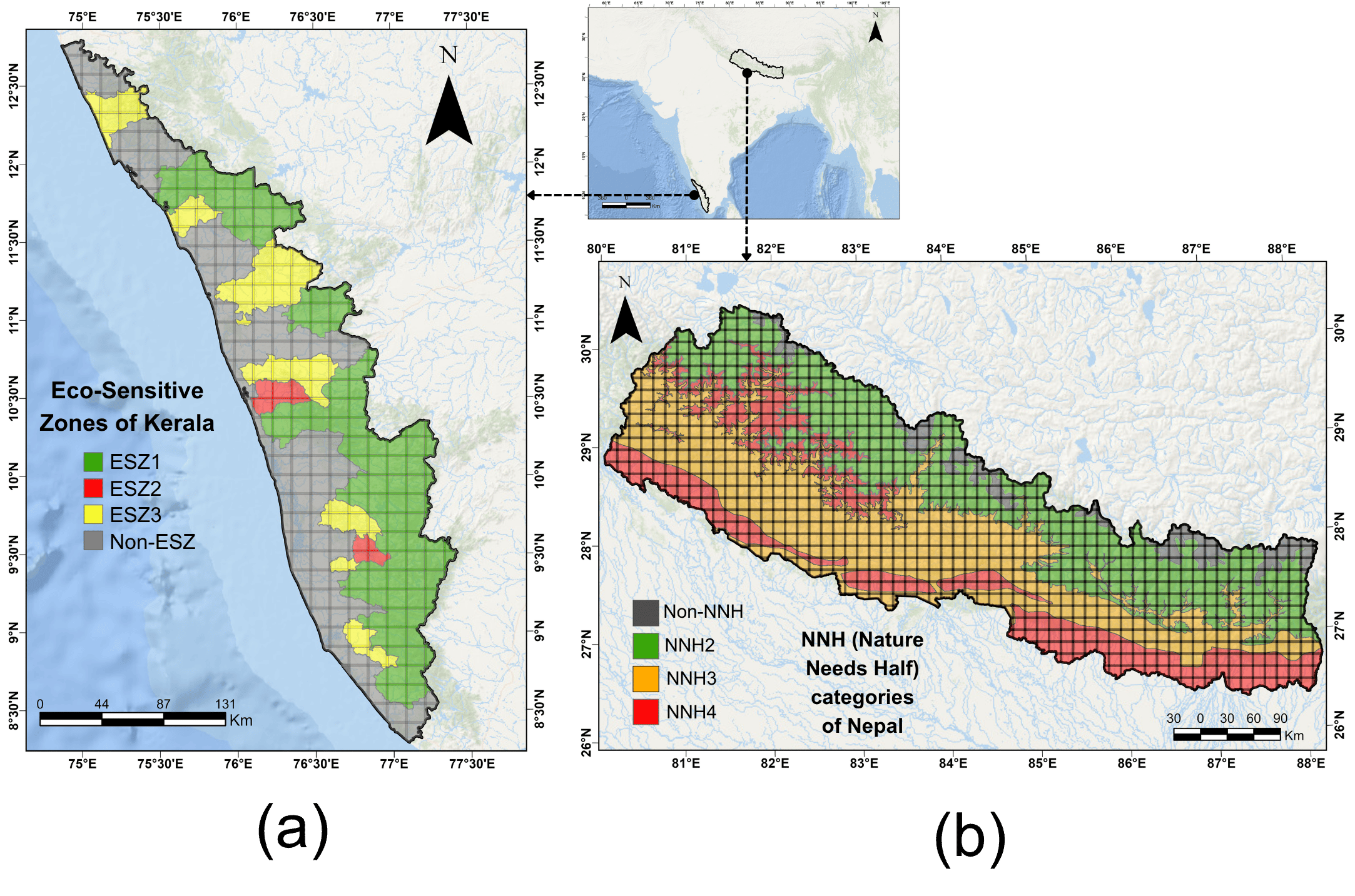}
    \caption{(Left) Schematic diagram of  Early Fusion (EF), Late Fusion (LF), and Mixture of Experts (MoE) using conditioning factors (CF) of hazards $H_1$ and $H_2$ to generate MHSM [$s_1$,$s_2$], and (Right) Two-level spatial representation of (a) Kerala and (b) Nepal using computational units and contextual zone maps.}
    \label{fig:spatialpartition}
\end{figure*}

In the proposed FL-MHSM workflow, spatial partitioning is carried out at two levels, \ie~for data partitioning and contextual zoning. This two-level strategy is applied consistently across the proposed models, \ie~the EF, LF, and MoE models. The partitioning of the study areas is shown in Figure~\ref{fig:spatialpartition}.

\begin{definition}[Computational Units]
Computational units are the partitions of the data for enabling parallelism, which, in the case of spatial data, use the additional constraint of contiguity and overlap for localized modeling and computation.
\end{definition}

At the first level, the region is subdivided into \(15\,\mathrm{km} \times 15\,\mathrm{km}\) computational units, \ie~lattice grids, with a \(1.5\,\mathrm{km}\) overlap on all sides. These units support localized susceptibility prediction, improve computational efficiency over large study areas, and help reduce boundary discontinuities during reconstruction of the final susceptibility maps.

At the second level, each study area is represented by a contextual zone map that defines the spatial contexts used for modeling across the relevant portions of the computational units. For instance, in our study areas, these correspond to ESZ-based zones together with Non-ESZ regions in Kerala, and to NNH-based zones together with Non-NNH regions in Nepal. Zone-specific conditioning factors are then applied within each zone, on all the first-level computational units contained in the zone. This ensures that the modeling reflects the regional environmental differences rather than enforcing a single feature configuration across the entire study area. 

In this strategy, computational units are not constrained to a single contextual zone. Units intersecting contextual-zone boundaries are handled through their zone-specific portions within the corresponding contextual zones.

This design is informed by our earlier large-area flood susceptibility mapping study, which showed that grid-based spatial organization supports scalable computation while preserving spatial continuity across neighboring units~\cite{mundayatt2024scaling}. Building on that finding, the present study introduces overlapping grids with contextual zone maps to provide a generalized and efficient spatially-adaptive workflow for EF, LF, and MoE modeling.

\subsection{Hazard Inventory Augmentation} \label{sec:hazardinvent}
The flood and landslide inventory maps from external sources are augmented separately for each hazard over the entire study area to support model development. 

For each hazard, to increase the count of positive samples, kernel density estimation (KDE) with a Gaussian kernel is applied to the observed hazard points~\cite{fan2023comparison}. Since the source inventories showed uneven spatial distributions, KDE helps to estimate the local probability surface of hazard occurrence from the spatial distribution of hazard coordinates~\cite{li2025evaluation}. Additional hazard samples are then drawn from the resulting density surface while preserving the local clustering characteristics of the original inventory. The upsampled points are further filtered using a minimum spacing constraint of 100~m to avoid oversampling at any site in the study areas.

Corresponding non-hazard inventories are generated separately for each hazard, \ie~yielding negative samples of the hazard occurrence for the respective prediction task. These samples are generated using a One-Class Support Vector Machine (OC-SVM) trained on the spatial coordinates of known hazard events~\cite{ye2023generating,chen2021one}. Random candidate points are sampled over the full study area and evaluated using the trained model to identify spatial outliers. Only those sites classified as ``non-hazard,'' \ie~non-occurrence of the concerned hazard, and located at least 100~m away from the known hazard points are retained from the candidate points.

A minimum spacing threshold of 100~m is adopted to match the working spatial resolution of the study, so that augmented hazard samples are not duplicated within the same effective spatial neighborhood and non-hazard samples remain separated from known hazard locations at the same analysis scale. These steps produce hazard-wise positive and negative samples over the full study area, which are subsequently used for training the zone-wise models.

\subsection{Zone-wise Feature Selection} \label{sec:feature_selection}
Feature selection is conducted independently within each contextual zone using a two-step procedure: the first step performs pairwise multicollinearity screening, and the second step determines feature-wise importance.

In the first step, linear association among continuous predictors is evaluated using the Pearson correlation coefficient, $r$~\cite{yousefi2020assessing}. For two variables $X_i$ and $X_j$, the coefficient is:
\begin{equation}
r_{ij} =
\frac{\sum_{k=1}^{n} (x_{k,i} - \bar{x}_i)(x_{k,j} - \bar{x}_j)}
{\sqrt{\sum_{k=1}^{n} (x_{k,i} - \bar{x}_i)^2}
\sqrt{\sum_{k=1}^{n} (x_{k,j} - \bar{x}_j)^2}},
\end{equation}
where $x_{k,i}$ denotes the value of feature $i$ for sample $k$, $\bar{x}_i$ is the mean of feature $i$, and $n$ is the number of observations in the corresponding zone. Pairwise correlations are examined, and when $|r_{ij}| > 0.8$, one of the correlated variables is removed to reduce redundancy. An alternative multicollinearity test is the Variance Inflation Factor (VIF)~\cite{o2007caution}, which provides a more comprehensive assessment. However, computing correlation is more efficient for a first-level screening to identify strong pairwise linear associations among features.

In the second step, feature importance is evaluated using SHapley Additive exPlanations (SHAP)~\cite{lundberg2017unified}. SHAP values are computed separately for each hazard-specific susceptibility model using an efficient tree-based estimator for feature attribution, \eg, LightGBM~\cite{al2025shapley,inan2023explainable}. For a trained model, SHAP decomposes $f(x)$, which is the prediction for a sample $x$, into feature-wise additive contributions of the $p$ input features:
\begin{equation}
f(x) = \phi_0 + \sum_{i=1}^{p} \phi_i(x),
\end{equation}
where $\phi_0$ is the expected prediction over the dataset and $\phi_i(x)$ denotes the contribution of the $i^{\text{th}}$ feature for sample $x$. For the $k^{\text{th}}$ sample, this contribution is written as $\phi_{k,i}$. Let $n$ denote the number of samples used to compute SHAP values. The global importance of feature $i$, denoted by $I_i$, was computed as the mean absolute SHAP value across all samples and then converted into percentage contribution $P_i$:
\begin{eqnarray}
I_i &=& \frac{1}{n} \sum\limits_{k=1}^{n} |\phi_{k,i}|, \\
P_i &=& \frac{I_i}{\sum\limits_{j=1}^{p} I_j} \times 100.
\end{eqnarray}

A feature is discarded from zone-wise modeling only if its SHAP contribution is zero for all hazards considered within that zone, \ie~both flood and landslide in our study areas. The resulting zone-specific feature sets are then used for subsequent modeling in both the EF and LF models, and more features are discarded in the later stages of the FL-MHSM workflow.

\subsection{Early Fusion Models} \label{sec:earlyfusion}
\begin{definition}[Early Fusion for MHSM]
A joint modeling strategy in which a single model learns the susceptibilities of multiple hazards simultaneously at each spatial location, either at the data or feature levels, is referred to as EF for MHSM.
\end{definition}

In this study, EF is implemented using a multilayer perceptron (MLP) trained with a multivariate Gaussian negative log-likelihood (MVG-NLL) loss~\cite{bishop2006pattern} to represent inter-hazard dependence and predictive uncertainty within a unified formulation~\cite{choi2025probabilistic,zanetta2025efficient}. The EF model is used in the MoE ensemble in the FL-MHSM workflow.

Let the target vector be \(y=[y_1,y_2,\ldots,y_D]^\top\), where \(y_i\) is the observed label for the $i$\textsuperscript{th} hazard for $D$ hazards, with label vector, \(H=[H_1,H_2,\ldots,H_D]\). The model assumes a multivariate normal distribution, given as:
\[
y \sim \mathcal{N}(\mu,\Sigma),
\]
where \(\mu=[\mu_1,\mu_2,\ldots,\mu_D]^\top\) is the predictive mean vector and \(\Sigma \in \mathbb{R}^{D\times D}\) is the predictive covariance matrix. Thus, the network must estimate both the expected susceptibility values and the covariance structure of $D$ hazards. Hereafter, we focus on the two-hazard case, \ie~$D$=2, thus using a bivariate normal distribution.

For the bivariate setting, the MLP outputs five values in total: two values correspond to the predictive means \((\mu_1,\mu_2)\), and three values correspond to the elements of a lower triangular matrix used to reconstruct the covariance matrix. Rather than predicting \(\Sigma\) directly, the covariance is decomposed into an inner product of a lower triangular matrix $L$ using Cholesky factorization to maintain positive definitiveness during training:\\
\centerline{$\Sigma = LL^\top$.}

For the two-hazard case,
\[
L =
\begin{bmatrix}
l_{11} & 0 \\
l_{21} & l_{22}
\end{bmatrix},
\quad
\Sigma =
\begin{bmatrix}
l_{11}^2 & l_{11}l_{21} \\
l_{11}l_{21} & l_{21}^2 + l_{22}^2
\end{bmatrix}.
\]

To maintain valid covariance parameters during optimization, the diagonal elements of \(L\) are predicted in log-space and exponentiated during reconstruction. The off-diagonal term of \(\Sigma\) captures the dependence between hazards $H_1$ and $H_2$ within the joint predictive distribution.

The MVG-NLL for a single observation is defined as:
\[
\mathcal{L}_{MVG} =
\tfrac{1}{2}\left[
D\log(2\pi) + \log|\Sigma| +
(y-\mu)^\top\Sigma^{-1}(y-\mu)
\right].
\]

This is rewritten using the Cholesky factorization by substituting $\log|\Sigma|=2\sum_{i=1}^{D}\log l_{ii}$,
\begin{equation}\mathcal{L}_{MVG}=\tfrac{D}{2}\log(2\pi) + \sum_{i=1}^{D}\log l_{ii} + \tfrac{1}{2}\|L^{-1}(y-\mu)\|^2.\end{equation}

From the reconstructed covariance matrix, inter-hazard correlation is computed as:\\
\centerline{$\rho = \tfrac{\sigma_{12}}{\sigma_1 \sigma_2}$,}\\
where \(\sigma_{12}\) is the covariance between the hazards $H_1$ and $H_2$, and \(\sigma_1\) and \(\sigma_2\) are the corresponding marginal standard deviations of $H_1$ and $H_2$, respectively. Predictive uncertainty is represented by the log-determinant \(\log|\Sigma|\), which summarizes the overall spread of the joint predictive distribution. Larger values indicate greater joint predictive uncertainty, whereas smaller values indicate more confident predictions and are desirable.

During inference, a sigmoid transformation is applied to the predicted means to obtain the susceptibility probabilities for $H_1$ and $H_2$ in the interval \([0,1]\).

\subsection{Late Fusion Models} \label{sec:latefusion}
\begin{definition}[Late Fusion for MHSM]
A modeling strategy in which the susceptibilities of multiple hazards, that are learned separately using hazard-specific models, are combined, \ie~a decision-level fusion or concatenation, at each spatial location, is referred to as LF for MHSM.
\end{definition}

XGB is used as the baseline model for LF modeling in FL-MHSM. It is a gradient-boosted decision tree algorithm known for strong predictive performance, regularization, and computational efficiency in structured geospatial datasets. These characteristics make it well-suited for modeling nonlinear relationships between environmental conditioning factors and hazard occurrence~\cite{abedi2022flash}.

In our workflow, separate XGB models are trained for two hazards, $H_1$ and $H_2$, within each contextual zone to generate hazard-specific susceptibility maps. The resulting hazard-specific probability outputs are then combined after prediction to form the LF output vector, \([p^{\mathrm{LF}}_{H_1}, p^{\mathrm{LF}}_{H_2}]\), at each spatial location. This LF model is used in the MoE ensemble.

\subsection{Ensemble Modeling Using Mixture of Experts} \label{sec:moe}
In FL-MHSM, EF and LF models offer complementary advantages. EF learns the two hazards jointly and can capture shared susceptibility structure arising from inter-hazard dependence, whereas LF models each hazard independently and can better preserve hazard-specific discrimination. Integrating these models will combine their benefits. A hybrid MoE model~\cite {jacobs1991adaptive} is adopted to combine the two experts within a unified soft-gating architecture. Only the final susceptibility probabilities of the two hazards, $H_1$ and $H_2$, from both experts are passed to the gating network for expert weighting.

The gating network is implemented as a shallow MLP that operates on the concatenated expert probabilities,
\[
[p^{\text{LF}}_{H_1},\, p^{\text{LF}}_{H_2},\, p^{\text{EF}}_{H_1},\, p^{\text{EF}}_{H_2}],
\]
where $p^{\text{LF}}_{H_1}$ and $p^{\text{LF}}_{H_2}$ denote the probabilities of $H_1$ and $H_2$ predicted by the LF expert, and $p^{\text{EF}}_{H_1}$ and $p^{\text{EF}}_{H_2}$ represent the corresponding probabilities from the EF expert. The gating network produces soft mixture weights through a softmax layer,
\[
[w_{\text{LF}}, w_{\text{EF}}] = \text{softmax}(g(z)),
\]
where $z$ denotes the concatenated expert probabilities and $g(\cdot)$ represents the gating MLP.

Final $H_1$ and $H_2$ susceptibility predictions are obtained through a weighted combination of the expert outputs,
\[
p^{\text{MoE}} = w_{\text{LF}}\,p^{\text{LF}} + w_{\text{EF}}\,p^{\text{EF}}.
\]

The MoE is trained using binary cross-entropy loss, allowing the gating network to learn how to combine independent and joint susceptibility patterns across spatial contexts.

\subsection{Spatial Heterogeneity Analysis of MoE MHSM} \label{sec:spatialhetero}
To investigate the spatial explanatory power of conditioning factors and their interactions on modeled susceptibility patterns, GeoDetector is applied to each contextual zone for each hazard in a study area. As the MoE model represents a soft-gated integration of hazard-specific LF predictions and jointly learned EF predictions, the analysis is performed on the MoE susceptibility outputs to examine how conditioning factors explain the spatial heterogeneity of the resulting fused susceptibility structure.

GeoDetector~\cite{wang2024statistical} is a statistical framework designed to quantify spatial stratified heterogeneity and identify the driving mechanisms underlying spatially distributed phenomena without requiring assumptions of linear relationships or distributional normality between conditioning factors and the response variable. The framework has been increasingly applied in landslide susceptibility studies to analyze spatial heterogeneity and interactions among environmental conditioning factors~\cite{yao2025heterogeneous,li2024enhancing}. As a preprocessing step, continuous conditioning factors were discretized into five quantile-based strata to satisfy the stratified input requirement of GeoDetector. The analysis consists of three sequential components: factor detection, interaction detection, and risk detection~\cite{wang2010geographical}.

\subsubsection{Factor Detector}
The explanatory power of an independent variable $X$ on the spatial distribution of susceptibility $Y$ is quantified using the $q$-statistic:
\begin{equation}
q = 1 - \frac{\sum_{h=1}^{L} N_h \sigma_h^2}{N \sigma^2},
\end{equation}
where $h$ denotes the stratum of factor $X$, $L$ is the total number of strata, $N_h$ and $\sigma_h^2$ represent the sample size and variance of $Y$ within stratum $h$, and $N$ and $\sigma^2$ denote the total sample size and global variance of $Y$, respectively.

The $q$-statistic ranges from 0 to 1, where larger $q$ values indicate stronger explanatory power of factor $X$ for the spatial distribution of susceptibility. In this study, $Y$ represents the modeled susceptibility values produced by the MoE model, while the conditioning factors serve as the explanatory variables $X$, allowing GeoDetector to evaluate how these factors explain the spatial distribution of modeled susceptibility rather than only the observed hazard distribution.

\subsubsection{Interaction Detector}
To examine whether the combined influence of two variables enhances, weakens, or acts independently in explaining susceptibility, an interaction value is computed. For two variables $X_1$ and $X_2$, the explanatory power of their interaction is calculated as:\\
\centerline{$q(X_1 \cap X_2)$.}

This value is compared with the individual explanatory powers $q(X_1)$ and $q(X_2)$ to determine the interaction type. Enhancement occurs under the following conditions:
\begin{eqnarray}
\text{Bi-factor: } &q(X_1 \cap X_2) &> \max\big(q(X_1), q(X_2)\big), \nonumber\\
\text{Nonlinear: } &q(X_1 \cap X_2) &> q(X_1) + q(X_2).\nonumber
\end{eqnarray}

This analysis reveals compound effects among conditioning factors that cannot be identified through single-factor assessment alone. It is particularly relevant in multi-hazard environments, where susceptibility is often shaped by combined climatic, hydrological, and topographic influences.

\subsubsection{Risk Detector}
The risk detector evaluates the directional influence of conditioning factors by comparing mean susceptibility values across the strata of each variable. For each factor, the mean susceptibility within each stratum is computed and statistically compared using one-way analysis of variance (ANOVA) to test whether susceptibility values differ significantly among factor strata. This analysis provides process-level interpretation by revealing how susceptibility varies along environmental gradients.

\subsection{Evaluation Metrics} \label{sec:eval}
The EF model, the baseline LF model, and their integrated MoE model can be alternatively used in FL-MHSM. For the optimal choice, model performance is assessed using classification, probabilistic, distributional, and spatial agreement metrics~\cite{pugliese2024hazard}. All the metrics selected for our work are bounded in the range of [0,1]. Let \(TP\), \(TN\), \(FP\), and \(FN\) denote true positives, true negatives, false positives, and false negatives, respectively, for \(N\) samples. 

\begin{itemize}

\item \textit{Accuracy:} measures the proportion of correctly classified samples among all observations. Higher values indicate better overall classification performance.
\begin{equation}
\mathrm{Accuracy} = \frac{TP + TN}{TP + TN + FP + FN}
\end{equation}

\item \textit{Precision:} is the proportion of predicted positive samples that correspond to true hazard occurrences. Higher values indicate fewer false positive predictions.
\begin{equation}
\mathrm{Precision} = \frac{TP}{TP + FP}
\end{equation}

\item \textit{Recall (True Positive Rate):} measures the proportion of actual positive samples correctly detected by the model. Higher recall implies more correct predictions of actual hazard occurrences.
\begin{equation}
\mathrm{Recall} = \frac{TP}{TP + FN}
\end{equation}

\item \textit{F\textsubscript{1}-score:} represents the harmonic mean of precision and recall, balancing false positives and false negatives. Higher values indicate a better precision-recall balance.
\begin{equation}
\mathrm{F}_1 = \frac{2TP}{2TP + FP + FN}
\end{equation}

\item \textit{AUC--ROC:} combines two metrics, namely,\\
(i) the Area Under the Curve (AUC) is the overall ability of the model to discriminate between hazard and non-hazard samples across all classification thresholds;\\
(ii) the Receiver Operating Characteristic (ROC) curve evaluates threshold-independent discrimination by plotting the true positive rate against the false positive rate,
\begin{equation}
\mathrm{FPR} = \frac{FP}{FP + TN}
\end{equation}
An AUC-ROC value of 0.5 indicates random discrimination, above 0.5 indicates better than random performance, and closer to 1 indicates stronger class separability.

\item \textit{Brier Score:} measures the mean squared difference between predicted probabilities and observed binary outcomes, thereby assessing probabilistic accuracy and calibration~\cite{liang2025evaluation}. Lower values imply more accurate prediction outcomes, and 0 implies perfect prediction.
\begin{equation}
\mathrm{Brier} = \frac{1}{N} \sum_{i=1}^{N} (p_i - y_i)^2
\end{equation}

\item \textit{Multi-Hazard Inventory Distribution (MH Density):} shows the distribution of observed hazard occurrences in joint susceptibility classes for two hazards, as the percentage of inventory points falling within each bivariate class.
\begin{equation}
\mathrm{MH}_{i,j} = \frac{N_{i,j}}{\sum_{i=1}^{S_1} \sum_{j=1}^{S_2} N_{i,j}} \times 100,
\end{equation}
where $N_{i,j}$ denotes the number of observed hazard occurrence points in the $(i,j)$ bivariate susceptibility cell, \ie~$i$ and $j$ correspond to the susceptibility level of hazards $H_1$ and $H_2$ for $S_1$ and $S_2$ severity levels, respectively. MH density is directly proportional to the concentration of hazard events within the corresponding joint susceptibility class. In our work, $S_1=S_2=5$, for both study areas.
 
\item \textit{Jaccard Index:} is used to quantify spatial agreement among susceptibility models in identifying critical hazard zones, and is computed on binary risk masks derived from the susceptibility maps~\cite{gharakhanlou2023flood,zhang2022earthquake}. For two binary maps $A$ and $B$, Jaccard similarity is defined as:
\begin{equation}
J(A,B) = \frac{|A \cap B|}{|A \cup B|},
\end{equation}
where $A \cap B$ denotes the number of spatial pixels classified as critical by both models, and $A \cup B$ denotes the number classified as critical by at least one model. Absence of spatial overlap and perfect spatial agreement give values 0 and 1, respectively.

\end{itemize}

\section{Experimental Setup and Implementation Details} \label{sec:expt}
All FL-MHSM experiments were conducted using the spatial partitioning strategy (Section~\ref{sec:spatialpart}). For both Kerala and Nepal, model training and susceptibility prediction were carried out separately within each contextual zone and applied consistently across the EF, LF, and MoE models. 

\subsection{Model Design and Training}
Each zone-specific model was trained, validated, and tested using the samples associated with the corresponding contextual zone. Hence, within each contextual zone, the dataset was partitioned using a stratified 80:10:10 split for training, validation, and testing, respectively, to preserve the class distribution of hazard occurrence across all subsets. Although hazard inventory augmentation and non-hazard sample generation were performed for the study area as a whole, the subsequent contextual-zone partitioning still results in class imbalance within individual zones. Therefore, macro-averaged AUC-ROC is used for model performance analysis for the entire study area to ensure balanced evaluation under zone-specific class-imbalanced settings.

For the LF model, the XGB baseline (Section~\ref{sec:latefusion}) was trained separately for flood and landslide susceptibility within each contextual zone. Hyperparameters were optimized using randomized search with three-fold cross-validation. The search space included the number of trees, maximum tree depth, learning rate, subsampling ratios, and class-weighting strategies.

For the proposed EF model, the MVG-MLP (Section~\ref{sec:earlyfusion}) network architecture consisted of three hidden layers with 256, 128, and 64 units, respectively, with Gaussian Error Linear Unit (GELU) activation, layer normalization, and a residual connection in the 256-dimensional block. Regularization was introduced through Gaussian noise (0.01) and dropout (0.10). Training was performed using the Adam optimizer with a fixed learning rate of $1 \times 10^{-4}$, a batch size of 256, and early stopping with a maximum of 100 epochs. During inference, the predicted means were transformed using a sigmoid function and subsequently calibrated using beta calibration to obtain final flood and landslide probabilities.

\begin{table}[tpb]
\centering
\caption{Configuration of the MVG-MLP-based EF Model.}
\label{tab:mvg_config}
\renewcommand{\arraystretch}{1.1}
\setlength{\tabcolsep}{4pt}

\begin{tabular}{@{}p{3.1cm} p{8.6cm}@{}}
\hline
\textbf{Component} & \textbf{MVG Model Specification} \\
\hline
Loss function & MVG-NLL \\
Outputs & $\mu_1, \mu_2$ + covariance parameters \\
Hidden layers & 256 -- 128 -- 64 \\
Activation function & GELU \\
Output activation & Linear \\
Normalization & Layer normalization \\
Residual connections & Yes (256-dim block) \\
Regularization & Gaussian noise (0.01), dropout (0.10) \\
Covariance modeling & Full covariance (Cholesky decomposition) \\
Optimizer & Adam \\
Learning rate & Fixed ($1 \times 10^{-4}$) \\
Batch size & 256 \\
Training epochs & Up to 100 (early stopping) \\
Probability transformation & Sigmoid applied to $\mu$ \\
Calibration method & Beta calibration \\
Calibrated outputs & Flood and landslide probabilities \\
\hline
\end{tabular}

\vspace{1mm}
\begin{minipage}{\linewidth}
\footnotesize
\textbf{Legend:} $\mu_1$ and $\mu_2$ denote the predicted mean susceptibility terms for flood and landslide, respectively.
\end{minipage}
\end{table}

For the soft-gating MoE model (Section~\ref{sec:moe}), the gating network was trained on the EF and LF probability outputs. The gating network consisted of two hidden layers with 32 and 16 units, using Rectified Linear Unit (ReLU) activation and dropout regularization of 0.20 and 0.10, respectively. The output layer used softmax activation to estimate the expert weights. The gating model was trained using binary cross-entropy loss with label smoothing of 0.05, the Adam optimizer, a learning rate of $1 \times 10^{-3}$, and a batch size of 256 for up to 200 epochs with early stopping.

The architectural settings of the MVG-based early-fusion model and the soft-gating MoE model are summarized in Tables~\ref{tab:mvg_config} and \ref{tab:moe_config}, respectively.

\begin{table}[tpb]
\centering
\caption{Configuration of the soft-gating MoE model.}
\label{tab:moe_config}
\renewcommand{\arraystretch}{1.1}
\setlength{\tabcolsep}{4pt}

\begin{tabular}{@{}p{3.2cm} p{8.5cm}@{}}
\hline
\textbf{Component} & \textbf{Setting} \\
\hline
Gating hidden layers & 32 -- 16 \\
Gating activation & ReLU \\
Gating regularization & \makecell[l]{Dropout \\ (0.20, 0.10)} \\
Gating output & \makecell[l]{Softmax weights \\ ($w_{\text{EF}}, w_{\text{LF}}$)} \\
Loss function & \makecell[l]{Binary cross-entropy \\ (label smoothing = 0.05)} \\
Optimizer & Adam \\
Learning rate & $1 \times 10^{-3}$ \\
Training epochs & \makecell[l]{Up to 200 \\ (early stopping)} \\
Batch size & 256 \\
\hline
\end{tabular}
\end{table}

\subsection{FL-MHSM Generation}
To enable susceptibility prediction across the full study areas, model inference was performed on \texttt{GeoPackage}~\cite{ogc_geopackage} tiles corresponding to the computational units generated during the spatial partitioning stage. Each tile represented an independent spatial unit containing the required conditioning-factor values. Prediction was conducted unit-wise within each contextual zone using zone-specific models.

In the XGB-based LF model, prediction across computational units was parallelized using the Python \texttt{Dask} framework, enabling data parallelism. For MVG-MLP-based EF, GPU-based batched inference was adopted, wherein zone-specific models were applied sequentially to the computational units. This scalable prediction strategy enabled efficient generation of large-scale multi-hazard susceptibility estimates while maintaining computational feasibility.

At each prediction location \(x\), the multi-hazard susceptibility output is represented as a \(D\)-dimensional vector \(\mathbf{s}(x) = [s_1(x), s_2(x), \ldots, s_D(x)]\), where \(D\) denotes the number of hazards and \(s_k(x)\) denotes the susceptibility associated with the \(k\)-th hazard; in this study, \(D=2\), corresponding to flood and landslide susceptibility.

After tile-level susceptibility estimates were generated for all prediction points, overlapping outputs were merged using Inverse Distance Weighting (IDW) to remove boundary artifacts and obtain continuous susceptibility probability maps for flood and landslide hazards. The hazard-specific rasters were then combined by value to form the $D$-dimensional susceptibility vector at each location.

The merged maps were then post-processed to derive bivariate susceptibility classes by discretizing the continuous susceptibility values. For each study area, global Jenks natural break thresholds were computed separately for the flood and landslide severity levels using sampled values from all merged maps. The severity range of each hazard susceptibility was then discretized into five ordinal classes or \textit{levels}, namely Very Low (VL), Low (L), Moderate (M), High (H), and Very High (VH), and the two classified hazard-specific layers were combined to generate a \(5 \times 5\) bivariate susceptibility grid. These bivariate outputs were used in the subsequent MH density and spatial agreement analyses (Sections~\ref{sec:eval} and~\ref{sec:spatialhetero}).

All experiments were implemented in Python, and system specifications are an AMD Ryzen 9 5950X 16-core processor with 32 GB RAM and NVIDIA RTX A4000 GPU. The LF baseline was developed using XGB, while the MVG-MLP-based EF model and the soft-gating MoE model were implemented in TensorFlow/Keras. Feature selection and interpretability analysis were performed using LightGBM and SHAP. Spatial preprocessing and raster--vector operations were carried out using NumPy, Pandas, GeoPandas, and Rasterio.

\section{Results} \label{sec:results}
Here, we present our experimental results with respect to feature selection outcomes and FL-MHSM implementation for floods and landslides in both study areas, Kerala and Nepal.

\subsection{Feature Selection Outcomes}
\begin{figure*}[htbp]
    \centering
    \includegraphics[width=0.95\textwidth]{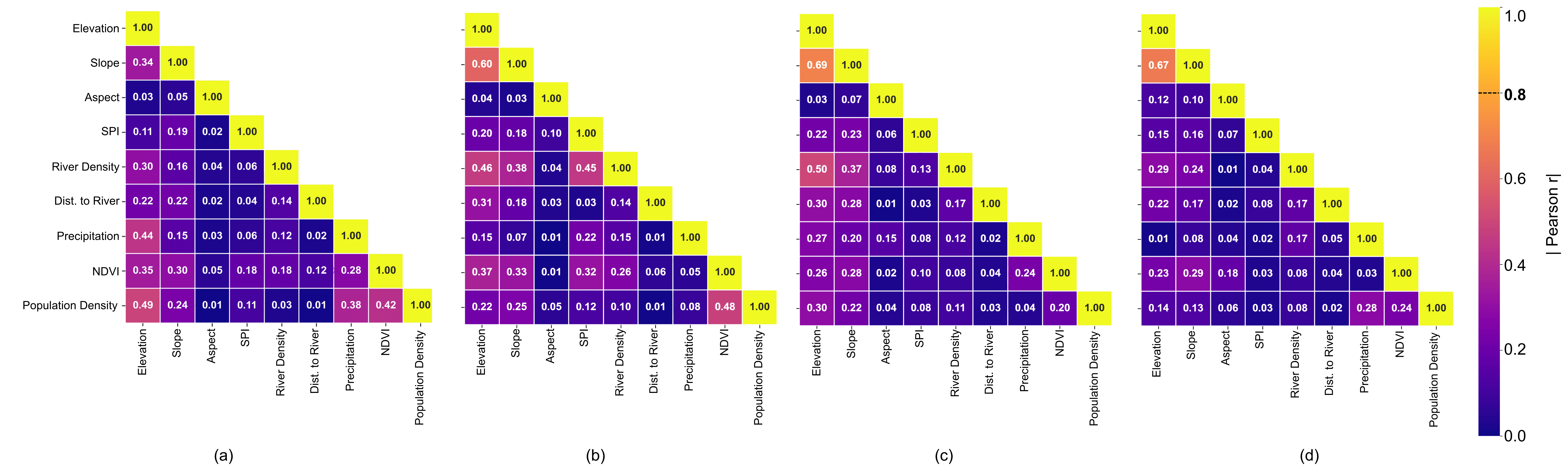}\\
    \small Correlation matrix for conditioning factors for Kerala (a) ESZ1, (b) ESZ2, (c) ESZ3, (d) Non-ESZ
    \\\vspace{1em}
    \includegraphics[width=0.9\textwidth]{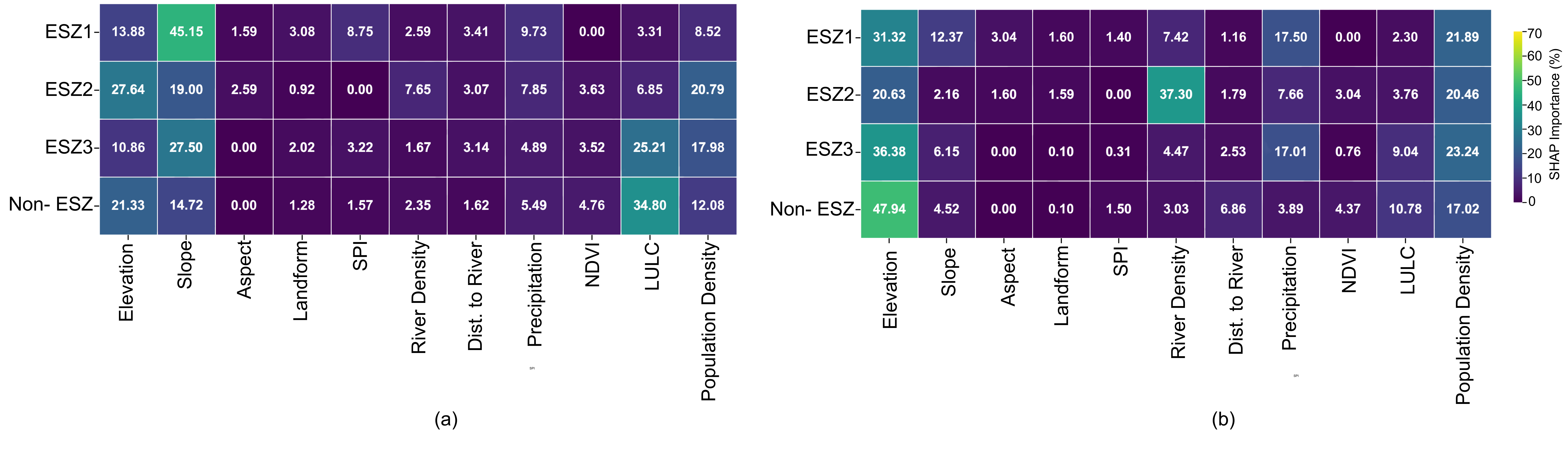}\\
    SHAP values for conditioning factors for Kerala  across ESZs for (a) Flood, (b) Landslide
    \caption{Feature selection from conditioning factors for the different contextual zones of Kerala using (Top) pairwise correlation analysis with a threshold of 0.8, and (Bottom) zone-wise SHAP values of all features.}
    \label{fig:featselect_kerala}

    \vspace{1em}
    
    \includegraphics[width=0.95\textwidth]{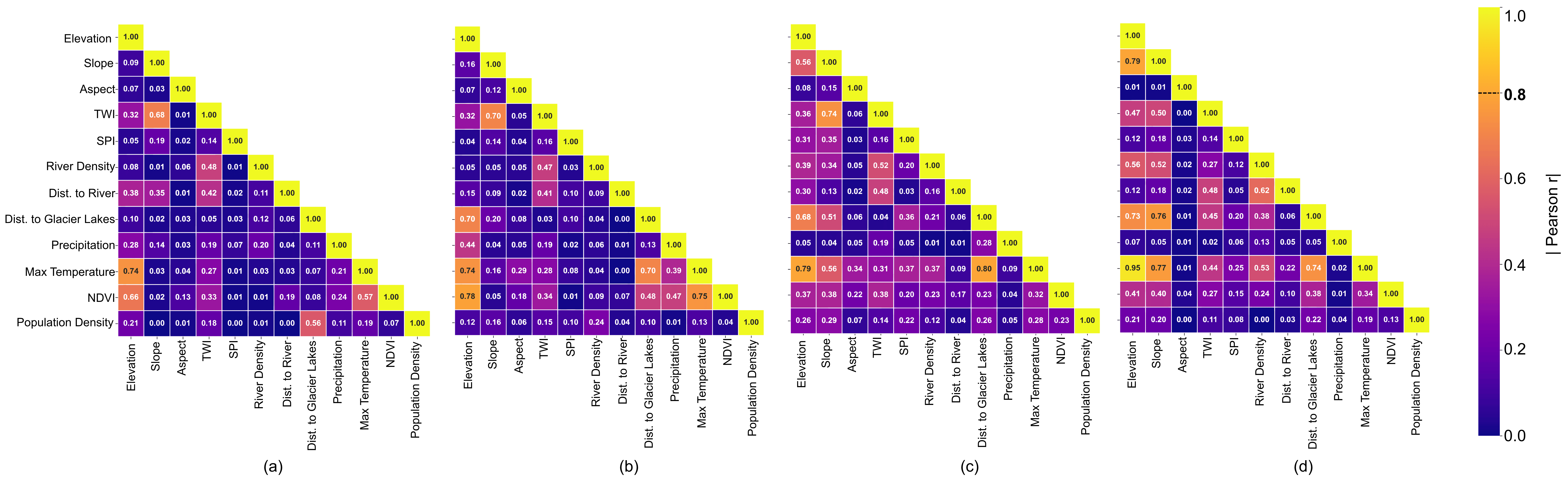}\\
    \small Correlation matrix for conditioning factors for Nepal (a) Non-NNH, (b) NNH2, (c) NNH3, (d) NNH4
    \\\vspace{1em}
    \includegraphics[width=0.9\textwidth]{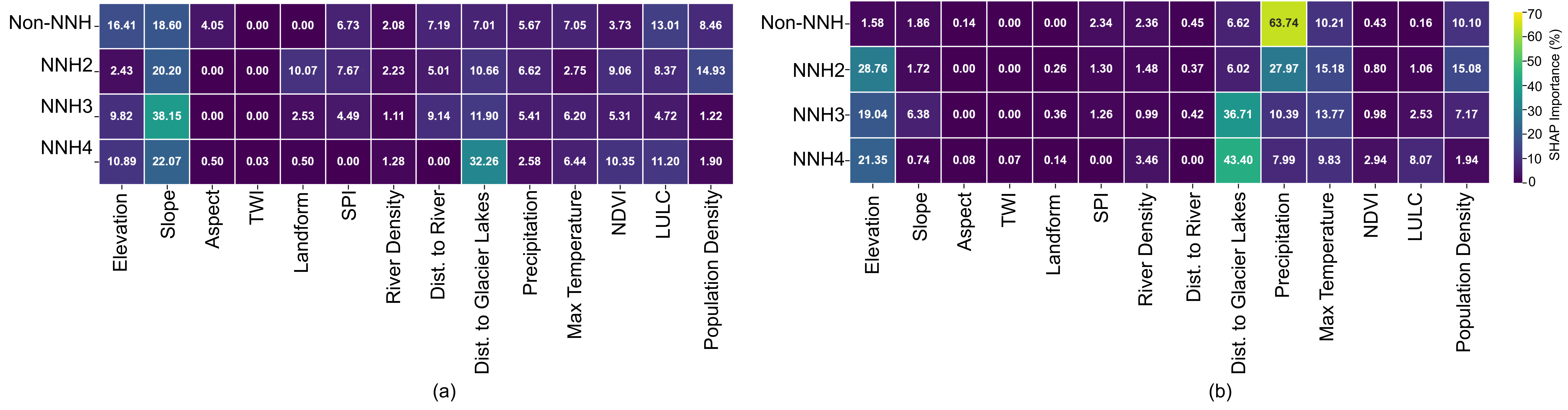}\\
    SHAP values for conditioning factors for Nepal across NNH zones for (a) Flood, (b) Landslide
    \caption{Feature selection from conditioning factors for the different contextual zones of Nepal using (Top) pairwise correlation analysis with a threshold of 0.8, and (Bottom) zone-wise SHAP values of all features.}
    \label{fig:featselect_nepal}
\end{figure*}

Zone-wise Pearson correlation screening and SHAP-based feature evaluation were applied to the conditioning factors for each hazard in each study area, using the procedure described in Section~\ref{sec:feature_selection}. The results for Kerala and Nepal are visualized in Figures~\ref{fig:featselect_kerala} and~\ref{fig:featselect_nepal}, respectively.

Across all zones, no predictor pairs exceeded the threshold of $|r| > 0.8$ for the correlation screening. Thus, in the absence of multicollinearity, all features were retained for SHAP-based assessment for both study areas. SHAP importance percentages were then computed separately for flood and landslide susceptibility. Predictors exhibiting zero SHAP contribution for both hazards within the same zone were excluded from the final feature subset. The SHAP-based filtering outcomes, as expected, show that retained predictors differ across the contextual zones for each hazard in each study area, and only the zone-relevant features are retained for subsequent modeling.

\subsubsection{Kerala}
NDVI in ESZ1, SPI in ESZ2, and Aspect in ESZ3 and Non-ESZ have been removed due to zero SHAP contribution. This shows that flood susceptibility in Kerala shows stronger terrain-related influence in ESZ1 and ESZ3, whereas landslide susceptibility exhibits more consistent elevation dominance across the zones. 

\subsubsection{Nepal}
TWI in Non-NNH, NNH2, and NNH3; Aspect in NNH2 and NNH3; Landform in Non-NNH; and Distance to River and SPI in NNH4 were removed due to zero SHAP contribution. This shows that: (i) landslide susceptibility is primarily precipitation-controlled in the lower zones, whereas glacier-lake proximity becomes more influential in the higher Himalayan zones, and (ii) flood susceptibility reflects a combination of terrain-related and cryosphere-related influences. 

The final set of features used for subsequent modeling is summarized in Table~\ref{tab:selected_factors_all}.

\begin{table}[t]
\centering
\caption{Final set of selected conditioning factors across contextual zones in Kerala and Nepal.}
\label{tab:selected_factors_all}
\setlength{\tabcolsep}{4pt}
\renewcommand{\arraystretch}{1.1}

\begin{adjustbox}{width=\linewidth}
\begin{tabular}{llcccc|cccc}
\toprule
\textbf{Category} & \textbf{Conditioning factor} & \textbf{ESZ1} & \textbf{ESZ2} & \textbf{ESZ3} & \textbf{Non-ESZ} & \textbf{Non-NNH} & \textbf{NNH2} & \textbf{NNH3} & \textbf{NNH4} \\
\midrule

\multirow{6}{*}{Topographic}
& Elevation  & \checkmark & \checkmark & \checkmark & \checkmark & \checkmark & \checkmark & \checkmark & \checkmark \\
& Slope      & \checkmark & \checkmark & \checkmark & \checkmark & \checkmark & \checkmark & \checkmark & \checkmark \\
& Aspect     & \checkmark & \checkmark & -         & -         & \checkmark & -         & -         & \checkmark \\
& TWI        & $\times$   & $\times$   & $\times$   & $\times$   & -         & -         & -         & \checkmark \\
& Landform   & \checkmark & \checkmark & \checkmark & \checkmark & -         & \checkmark & \checkmark & \checkmark \\
& SPI        & \checkmark & -         & \checkmark & \checkmark & \checkmark & \checkmark & \checkmark & - \\
\midrule

\multirow{3}{*}{Hydrological}
& River Density             & \checkmark & \checkmark & \checkmark & \checkmark & \checkmark & \checkmark & \checkmark & \checkmark \\
& Distance to River         & \checkmark & \checkmark & \checkmark & \checkmark & \checkmark & \checkmark & \checkmark & - \\
& Distance to Glacier Lakes & $\times$   & $\times$   & $\times$   & $\times$   & \checkmark & \checkmark & \checkmark & \checkmark \\
\midrule

\multirow{2}{*}{Climatic}
& Precipitation   & \checkmark & \checkmark & \checkmark & \checkmark & \checkmark & \checkmark & \checkmark & \checkmark \\
& Max Temperature & $\times$   & $\times$   & $\times$   & $\times$   & \checkmark & \checkmark & \checkmark & \checkmark \\
\midrule

\multirow{2}{*}{Surface Characteristics}
& NDVI & -         & \checkmark & \checkmark & \checkmark & \checkmark & \checkmark & \checkmark & \checkmark \\
& LULC & \checkmark & \checkmark & \checkmark & \checkmark & \checkmark & \checkmark & \checkmark & \checkmark \\
\midrule

\multirow{1}{*}{Anthropogenic}
& Population & \checkmark & \checkmark & \checkmark & \checkmark & \checkmark & \checkmark & \checkmark & \checkmark \\
\bottomrule
\end{tabular}
\end{adjustbox}

\begin{flushleft}
\footnotesize $\checkmark$ indicates selected factors, - indicates factors considered but not selected, and $\times$ indicates factors not considered for that study area.
\end{flushleft}
\end{table}

\subsection{Model Performance Evaluation}
\begin{figure*}[htbp]
    \centering
    \includegraphics[width=\textwidth]{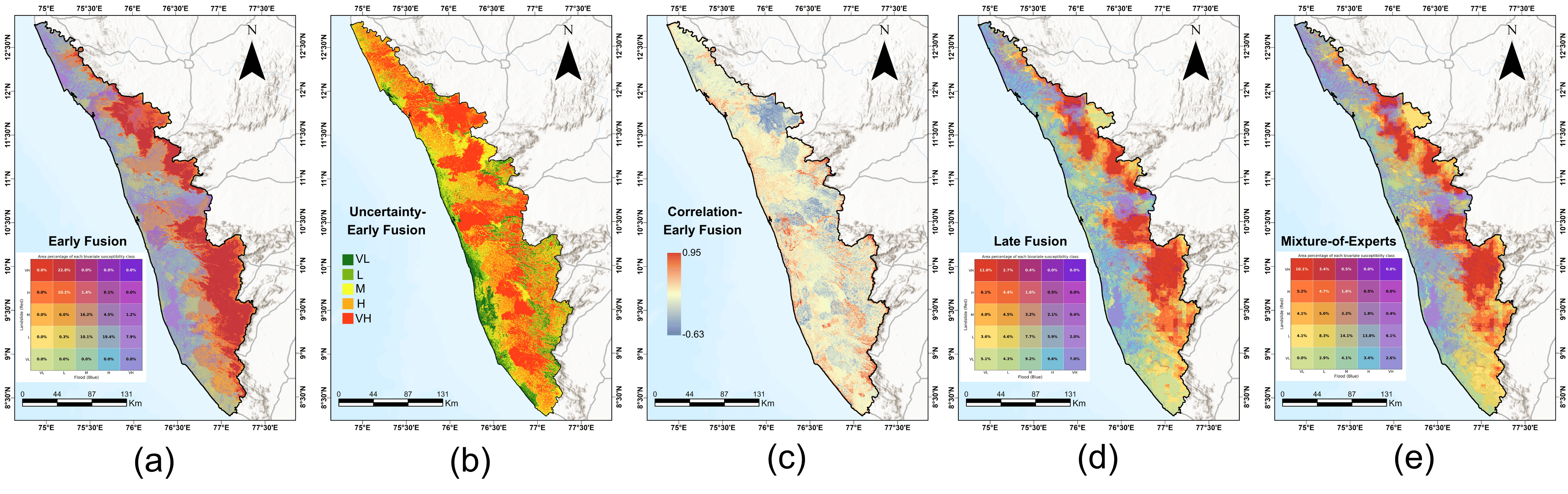}
     \caption{MHSM of Kerala using (a) EF, along with its uncertainty and correlation maps in (b) and (c), respectively; (d) LF; and (e) MoE, using a bivariate colormap to show the extent of area in each bivariate severity level of susceptibility value.}
    \label{fig:kerala_mhsm}

    \vspace{1em}
    
    \includegraphics[width=\textwidth]{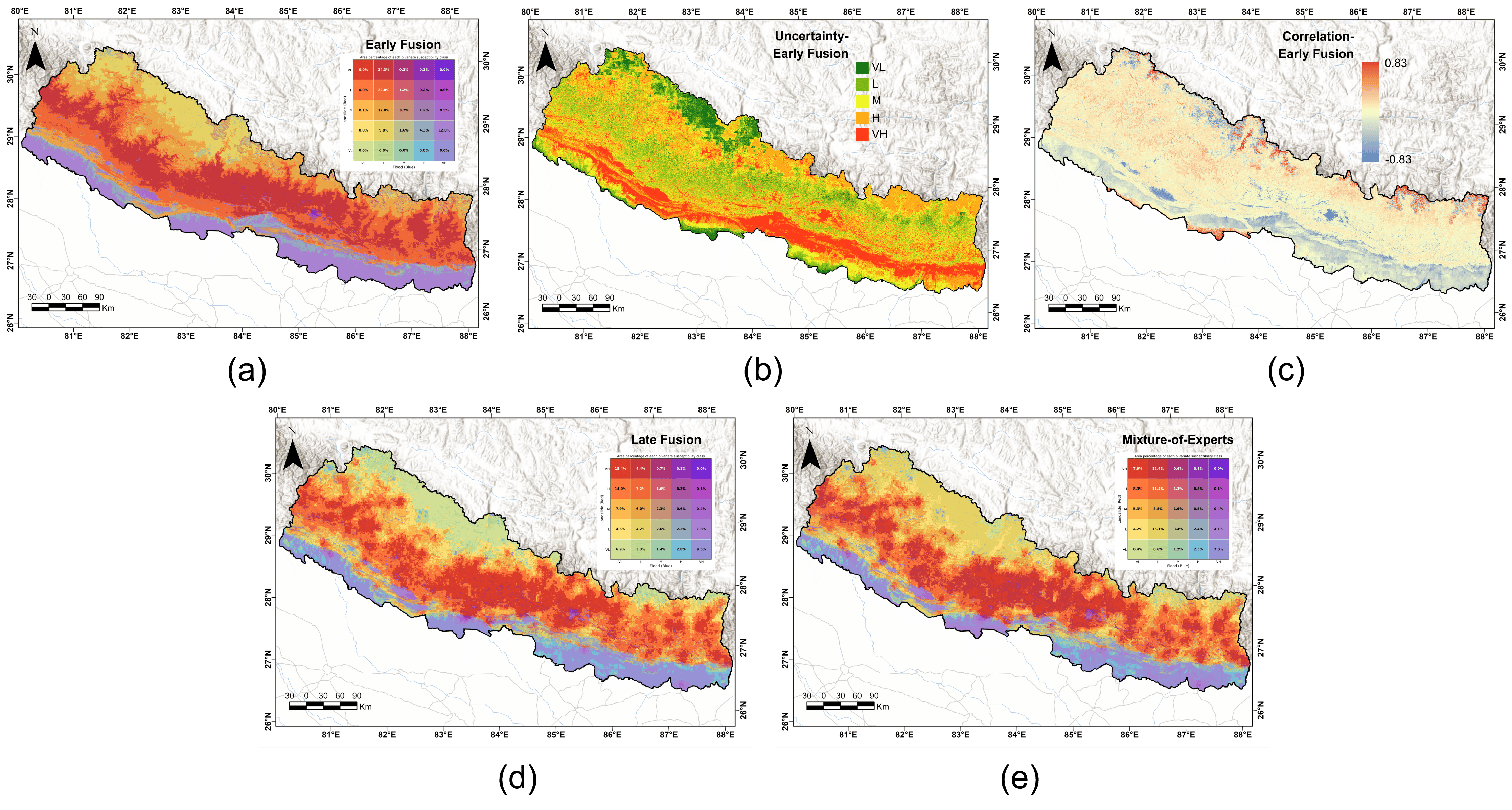}
     \caption{MHSM of Nepal using (a) EF, along with its uncertainty and correlation maps in (b) and (c), respectively; (d) LF; and (e) MoE, using a bivariate colormap to show the extent of area in each bivariate severity level of susceptibility value.}
    \label{fig:nepal_mhsm}
\end{figure*}

\begin{table}[htbp]
\centering
\caption{Macro-averaged performance of the fusion models, showing the best models for \{{\color{blue}\bf Flood}, {\color{red}\bf Landslide}\}.}
\label{tab:performance_combined}
\setlength{\tabcolsep}{4pt}
\renewcommand{\arraystretch}{1.05}

\begin{adjustbox}{width=\linewidth}
\begin{tabular}{lllcccccc}
\hline
\textbf{Region}
& \textbf{Model} & \textbf{Hazard}
& \textbf{AUC\textsubscript{ROC}} $\uparrow$ & \textbf{Precision} $\uparrow$
& \textbf{Recall} $\uparrow$ & \textbf{F\textsubscript{1}} $\uparrow$
& \textbf{Brier} $\downarrow$ & \textbf{Accuracy} $\uparrow$ \\
\hline

\multirow{6}{*}{Kerala}

& \multirow{2}{*}{Late Fusion (XGB)}
& Flood
& 0.898 & 0.575
& 0.816 & 0.675
& 0.092 & \color{blue}\bf 0.827 \\
&& Landslide
& \color{red}\bf 0.926 & 0.511
& 0.924 & 0.658
& \color{red}\bf 0.074 & 0.849 \\

& \multirow{2}{*}{Early Fusion (MLP-MVG)}
& Flood
& 0.875 & 0.565
& 0.840 & 0.676
& 0.086 & 0.780 \\
&& Landslide
& 0.880 & 0.480
& 0.860 & 0.616
& 0.088 & 0.732 \\

& \multirow{2}{*}{MoE}
& Flood
& \color{blue}\bf 0.905 & \color{blue}\bf 0.590
& \color{blue}\bf 0.930 & \color{blue}\bf 0.722
& \color{blue}\bf 0.082 & 0.824 \\
&& Landslide
& 0.924 & \color{red}\bf 0.520
& \color{red}\bf 0.940 & \color{red}\bf 0.670
& 0.078 & \color{red}\bf 0.864 \\

\hline

\multirow{6}{*}{Nepal}

& \multirow{2}{*}{Late Fusion (XGB)}
& Flood
& \color{blue}\bf 0.918 & 0.468
& 0.820 & 0.596
& 0.057 & 0.872 \\
&& Landslide
& 0.877 & 0.402
& 0.810 & 0.537
& 0.099 & 0.768 \\

& \multirow{2}{*}{Early Fusion (MLP-MVG)}
& Flood
& 0.900 & 0.445
& 0.858 & 0.586
& \color{blue}\bf 0.049 & 0.850 \\
&& Landslide
& 0.878 & 0.382
& 0.831 & 0.523
& 0.089 & 0.752 \\

& \multirow{2}{*}{MoE}
& Flood
& 0.913 & \color{blue}\bf 0.470
& \color{blue}\bf 0.898 & \color{blue}\bf 0.617
& 0.050 & \color{blue}\bf 0.875 \\
&& Landslide
& \color{red}\bf 0.914 & \color{red}\bf 0.405
& \color{red}\bf 0.901 & \color{red}\bf 0.559
& \color{red}\bf 0.088 & \color{red}\bf 0.770 \\

\hline 
\end{tabular}
\end{adjustbox}

\vspace{0.25em}
\begin{minipage}{\linewidth}
\textit{Note: The LF model performance measures are equivalent to those of running the flood and landslide susceptibility models separately.}
\end{minipage}
\end{table}

The visualization of outcomes of FL-MHSM in Kerala and Nepal is in Figures~\ref{fig:kerala_mhsm} and~\ref{fig:nepal_mhsm}, respectively. Evaluation metrics in Table~\ref{tab:performance_combined} were computed independently for each contextual zone and subsequently macro-averaged to obtain region-level performance estimates for each modeling strategy and hazard.

\subsubsection{Kerala} For flood susceptibility, EF gave higher Recall and Brier score compared to LF, while remaining comparable in F\textsubscript{1}-score. LF, however, gave a higher AUC-ROC and Accuracy. MoE performed strongest overall with the highest AUC-ROC, Precision, Recall, and F\textsubscript{1}-score, together with the lowest Brier score, with an Accuracy value comparable to LF.

For landslide susceptibility, EF underperformed compared to other models with lower AUC-ROC, Precision, Recall, F\textsubscript{1}-score, and Accuracy. LF gave higher AUC-ROC and Brier score, but MoE performed best on Precision, Recall, F\textsubscript{1}-score, and Accuracy, and performed comparable to LF in AUC-ROC and Brier score. 

\subsubsection{Nepal}
In flood susceptibility, EF improved the Recall and Brier score relative to LF, while LF retained the highest AUC-ROC. MoE recorded the highest Precision, Recall, F\textsubscript{1}-score, and Accuracy, and was comparable to LF in AUC-ROC and close to EF in Brier score. 

For landslide susceptibility, EF and LF were comparable in different respects, with EF improving the Recall and Brier score and LF remaining close in AUC-ROC. However, MoE outperformed across all evaluation metrics, including AUC-ROC, Precision, Recall, F\textsubscript{1}-score, Brier score, and Accuracy. 

Overall, the regional results show that EF improved the Recall and Brier score in several cases, while MoE achieved the most balanced performance across metrics by integrating EF and LF. LF was an optimal hazard-wise baseline, particularly for AUC-ROC and, in some cases, Brier score, but neither EF nor LF alone matched the consistency of MoE. These results show the overall efficacy of the proposed FL-MHSM workflow for large-scale multi-hazard susceptibility mapping.

\subsection{Spatial Agreement of High-Susceptibility Zones}
\begin{figure*}[!t]
    \centering
     \includegraphics[width=1\textwidth]{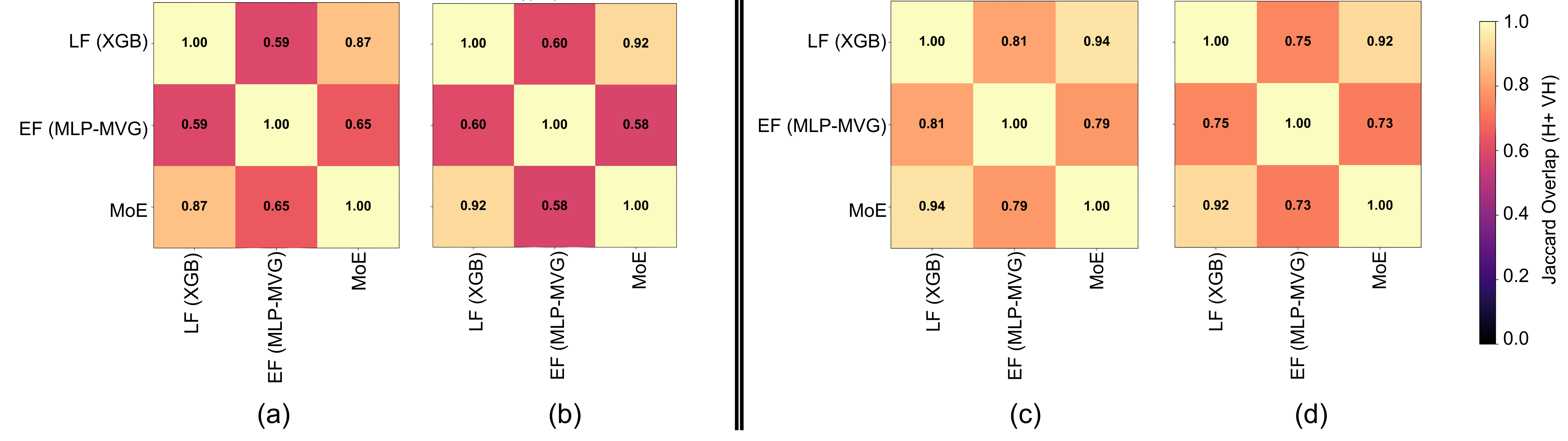}
    \caption{Jaccard Overlap of High--Very High Susceptibility Classes,  (a) Kerala - Flood, (b) Kerala - Landslide, (c) Nepal - Flood, (d) Nepal - Landslide}
    \label{fig:jaccardoverlap}
\end{figure*}

To assess structural consistency across fusion and ensemble strategies, spatial agreement was quantified using the Jaccard overlap index computed for the combined H--VH susceptibility classes, with respect to the five severity levels \{VL,L,M,H,VH\}. The comparative analysis of LF, EF, and the MoE models is shown as a heatmap in Figure~\ref{fig:jaccardoverlap}.

\subsubsection{Kerala}
In Kerala, structural differences among the three modeling strategies are more pronounced than in Nepal. For both hazards, the highest Jaccard overlap is observed between LF and MoE, indicating that the MoE outputs remain closest to the hazard-separated spatial pattern produced by LF. By contrast, the LF--EF overlap is substantially lower, and the EF--MoE agreement is only moderate, indicating a stronger structural shift under EF.

This pattern is evident for both hazards and is slightly stronger for landslides, where the EF--MoE overlap is the lowest among the Kerala model pairs. Overall, the Kerala results indicate that EF produces the largest redistribution of H--VH susceptibility zones for each hazard, while MoE moderates this shift and has outcomes closer to LF's results.

\subsubsection{Nepal}
In Nepal, the spatial agreement is relatively high across all model pairs. As in Kerala, the highest overlap is observed in MoE--LF for both hazards, showing high MoE--LF similarity in the delineation of H--VH susceptibility zones. However, higher LF--EF and EF--MoE overlaps indicate less intense structural effect of EF in Nepal than in Kerala.

Overall, among the two hazards, flood shows the highest overall agreement across models, whereas landslide shows visible divergence. Even so, similar outcomes are observed across both hazards, with MoE--LF showing the highest similarity and LF--EF with the highest dissimilarity. Results indicate greater structural stability in Nepal than in Kerala, with a smaller EF-driven redistribution of H--VH areas, \ie~areas with high susceptibility.

The Jaccard agreement results indicate that the structural effect of fusion is region-dependent, with stronger redistribution in Kerala and higher structural stability in Nepal. Although MoE remains structurally closer to LF than to EF in both regions, its improved predictive performance relative to LF suggests that the complementary information contributed by EF is useful even when the final spatial patterns do not fully shift toward the EF outcomes.

\subsection{Spatial Patterns and MH Density Validation of FL-MHSM} 
\begin{figure*}[!t]
    \centering
     \includegraphics[width=\textwidth]{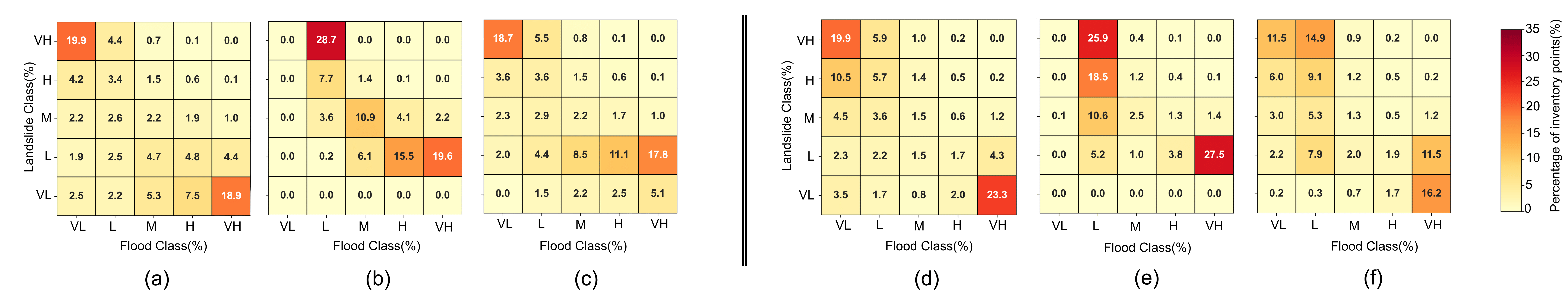}
     \caption{Heatmaps of inventory distribution (given as percentage of all training points) across 5$\times$5 bivariate susceptibility classes for LF (XGB), EF (MLP-MVG), and MoE in Kerala in ((a)-(c)) and Nepal in ((d)-(f)), respectively.}
    \label{fig:mhdensity}
\end{figure*}

The spatial distribution of bivariate susceptibility classes for Kerala and Nepal (in Figures~\ref{fig:kerala_mhsm} and~\ref{fig:nepal_mhsm}, respectively) is studied using the extent of multi-hazard patterns in the inventories, as shown in the heatmaps in Figure~\ref{fig:mhdensity}. We compared the performance of LF, EF, and MoE models for the spatial extent of multi-hazard susceptibility patterns. Using severity levels \{VL,L,M,H,VH\}, we identify regions in the compound susceptibility classes for flood-landslide hazards. Note that an \textit{X-Y} flood-landslide susceptibility class should be read as \textit{X} severity-level of flood susceptibility compounded with \textit{Y} severity-level of landslide susceptibility.

\subsubsection{Kerala}
In EF, Kerala MHSM is dominated by L--VH flood-landslide and other intermediate compound classes. L--VH and L--H flood-landslide classes together cover one-third of the study area, while M--M classes and H--L flood-landslide also occupy a substantial portion of the region, forming broad transitional belts between the coastal plains and the Western Ghats. In the corresponding heatmap, L--VH and VH--L flood-landslide are the dominant classes, while M--M density reaches 10.9\%. There are no occurrences of VL--VL inventory.

Compared with the EF pattern, LF preserves stronger hazard segregation along the coastal-highland gradient. Landslide-dominant classes are concentrated toward the Western Ghats, with VL--VH and VL--H flood-landslide classes together exceeding 17\% of the region, while flood-dominant classes are concentrated in the coastal plains, with H--VL and VH--VL flood-landslide classes together giving 16.6\%. Moderate compound classes map to transitional midland zones, but the H--H class is minimal at 0.5\%. Inventory density likewise concentrates in very high-very low severity combinations, particularly VL--VH and VH--VL flood-landslide classes, while the opposite diagonal classes are sparse.

The MoE model shows the main coastal-highland organization, also seen in LF, while moderating its stronger hazard segregation and incorporating part of the compound-class redistribution observed under EF. VL--VH flood--landslide susceptibility accounts for 10.1\%, while compound low-to-moderate classes are substantial, including M--L (14.1\%) and H--L (13.8\%) classes, producing smoother coastal-highland transitions than LF. In the heatmap, VL--VH flood-landslide (18.7\%) is significant, while flood-dominant classes such as VH--L class (17.8\%) are significant.

\subsubsection{Nepal}
Under the EF, the Nepal MHSM is dominated by VH--L flood-landslide and intermediate compound classes distributed across the Himalayan belt. L--VH and L--H flood-landslide classes together account for nearly 47\% of the total area. Flood-only extreme classes are largely suppressed, while moderate compound classes, such as L--M and M--M flood-landslide classes, become more prominent, indicating smoother transitions across susceptibility states. In the corresponding heatmap, inventory points cluster in L--VH and L--H flood-landslide classes, while the opposite-corner dominance declines relative to LF. Event density is therefore distributed more gradually across adjacent compound classes.

Compared with the EF outcomes, the LF model preserves stronger hazard segregation along Nepal's physiographic gradient. Landslide-dominant classes concentrate in the mountainous north, with VL--VH and VL--H flood-landslide classes together accounting for nearly 30\% of the total area. In contrast, the southern Terai exhibits flood-dominant classes, particularly the VH--VL flood-landslide class. Moderate combinations, such as VL--M  and L--M flood-landslide classes, define transitional belts across the Middle Mountains. True compound H--H classes are negligible ($<0.5\%$), indicating limited spatial overlap of extreme susceptibility states. The heatmap confirms strong segregation, with inventory points concentrating in opposite-corner combinations, especially VL--VH and VH--VL flood-landslide classes, exceeding 43\% of total occurrences. Diagonal H--H classes are sparse, reflecting hazard decoupling under independent modeling.

The MoE model shows the broad physiographic separation, as seen in LF, while moderating its stronger opposite-corner dominance and incorporating part of the smoother compound-class redistribution observed under EF. Landslide-dominant combinations are present but are moderated relative to EF, with L--VH and L--H flood-landslide classes covering 23.8\% of the area. Flood-dominant classes partially re-emerge in the southern plains, including VH--VL flood-landslide classes. Moderate combinations are also more evenly distributed, including L--L, L--H, and VH--L (11.5\%) flood-landslide classes. The H--H classes are limited (0.3\%). The heatmap confirms this moderated segregation: opposite-corner concentrations persist, particularly VH--VL and L--VH flood-landslide classes, but are reduced relative to LF, while distribution across neighboring classes becomes more gradual.

Overall, the regional patterns indicate that the structural effect of fusion is region-dependent. In both study areas, EF produces broader mixed-hazard distributions, LF preserves stronger hazard separation, and MoE gives an intermediate balanced outcome, with these full class-distribution contrasts appearing more pronounced in Nepal.

\subsection{Uncertainty and Inter-Hazard Correlation in EF Model}
Spatial variation in joint predictive uncertainty and inter-hazard correlation was examined for the EF-MHSM outputs using the zonal statistics reported in Table~\ref{tab:uncert_corr_zonal}, together with the uncertainty and correlation maps shown in Figures~\ref{fig:kerala_mhsm} and~\ref{fig:nepal_mhsm}, to compare the EF behavior across Kerala and Nepal.

\subsubsection{Kerala}
Zonal statistics showed that joint log-uncertainty was highest in ESZ2 and lowest in Non-ESZ, with ESZ1 and ESZ3 lying in between. Inter-hazard correlation was negative across all zones, with lower absolute values in Non-ESZ and ESZ1, and higher values in ESZ2 and ESZ3. The uncertainty map showed lower-uncertainty regions as fragmented local patches, whereas higher uncertainty was distributed more broadly across the state. Across flood-landslide susceptibility classes with non-zero area percentage, the lowest uncertainty occurred in VH--L, followed by L--L and L--VH, and was relatively low in M--L and H--L. Uncertainty increased in L--H, L--M, and VH--M, and was comparatively higher in H--M and M--M, reaching its highest represented levels in M--H and H--H. The correlation map showed a predominantly negative spread, with stronger negative patches occurring locally within a broader weakly negative background. Across flood-landslide susceptibility classes with non-zero area percentage, correlation was negative in all represented classes. The strongest negative values occurred in H--H and M--H, followed by VH--M, while the remaining represented classes showed weaker negative correlation.

\subsubsection{Nepal}
Zonal statistics indicated that joint log-uncertainty was lowest in NNH2 and highest in Non-NNH, with NNH3 and NNH4 lying in between. Inter-hazard correlation was negative across all zones, with the lowest absolute value in Non-NNH and the highest value in NNH4, followed by NNH3 and NNH2. The uncertainty map showed broader and more continuous stretches of higher uncertainty, whereas lower-uncertainty areas occurred as localized belts and clustered patches. Across flood-landslide susceptibility classes with non-zero area percentage, the lowest uncertainty occurred in L--L, followed by L--VH and L--H, and was relatively low in VH--L and L--M. Uncertainty then increased through M--L and VL--M, and was comparatively higher in M--M, M--H, H--L, M--VH, VH--M, and H--M, reaching its highest represented levels in H--H and H--VH. The correlation map showed a predominantly weakly negative spread, with stronger negative belts and localized patches embedded within a broader near-neutral background. Across flood-landslide susceptibility classes with non-zero area percentage, correlation was negative in nearly all represented classes. The strongest negative values occurred in H--VH and H--H, followed by VH--M, M--VH, H--M, and M--H, while the remaining represented classes showed weaker negative correlation. VL--M was the only represented class showing a weak positive correlation.

\begin{table}[!t]
\centering
\caption{Zone-wise mean joint log-uncertainty and inter-hazard correlation for the EF-MHSM outputs in Kerala and Nepal}
\label{tab:uncert_corr_zonal}
\renewcommand{\arraystretch}{1.1}
\setlength{\tabcolsep}{5pt}

\begin{tabular}{@{}l l c c@{}}
\hline
\textbf{Study Area} & \textbf{Zone} & \(\mathbf{\overline{U}_{\log}}\) & \(\mathbf{\bar{\rho}}\) \\
\hline
\multirow{5}{*}{Kerala}
& ESZ1             & -6.0568 & -0.1152 \\
& ESZ2             & -3.6979 & -0.1406 \\
& ESZ3             & -5.0965 & -0.1452 \\
& Non-ESZ          & -6.8485 & -0.1091 \\
& Whole study area & -6.1467 & -0.1185 \\
\hline
\multirow{5}{*}{Nepal}
& Non-NNH             & -6.0114 & -0.0303 \\
& NNH2             & -7.3572 & -0.0707 \\
& NNH3             & -6.5800 & -0.0895 \\
& NNH4             & -6.4612 & -0.1283 \\
& Whole study area & -6.7886 & -0.0892 \\
\hline
\end{tabular}

\vspace{1mm}
\begin{minipage}{\linewidth}
\footnotesize
\(\overline{U}_{\log}\) denotes the zone-wise mean joint log-uncertainty, computed from \(\log|\Sigma|\), and \(\bar{\rho}\) denotes the zone-wise mean inter-hazard correlation.
\end{minipage}
\end{table}

\subsection{Spatial Heterogeneity Analysis of MoE Ensemble Model}
\begin{figure*}[!t]
    \centering
     \includegraphics[width=0.9\textwidth]{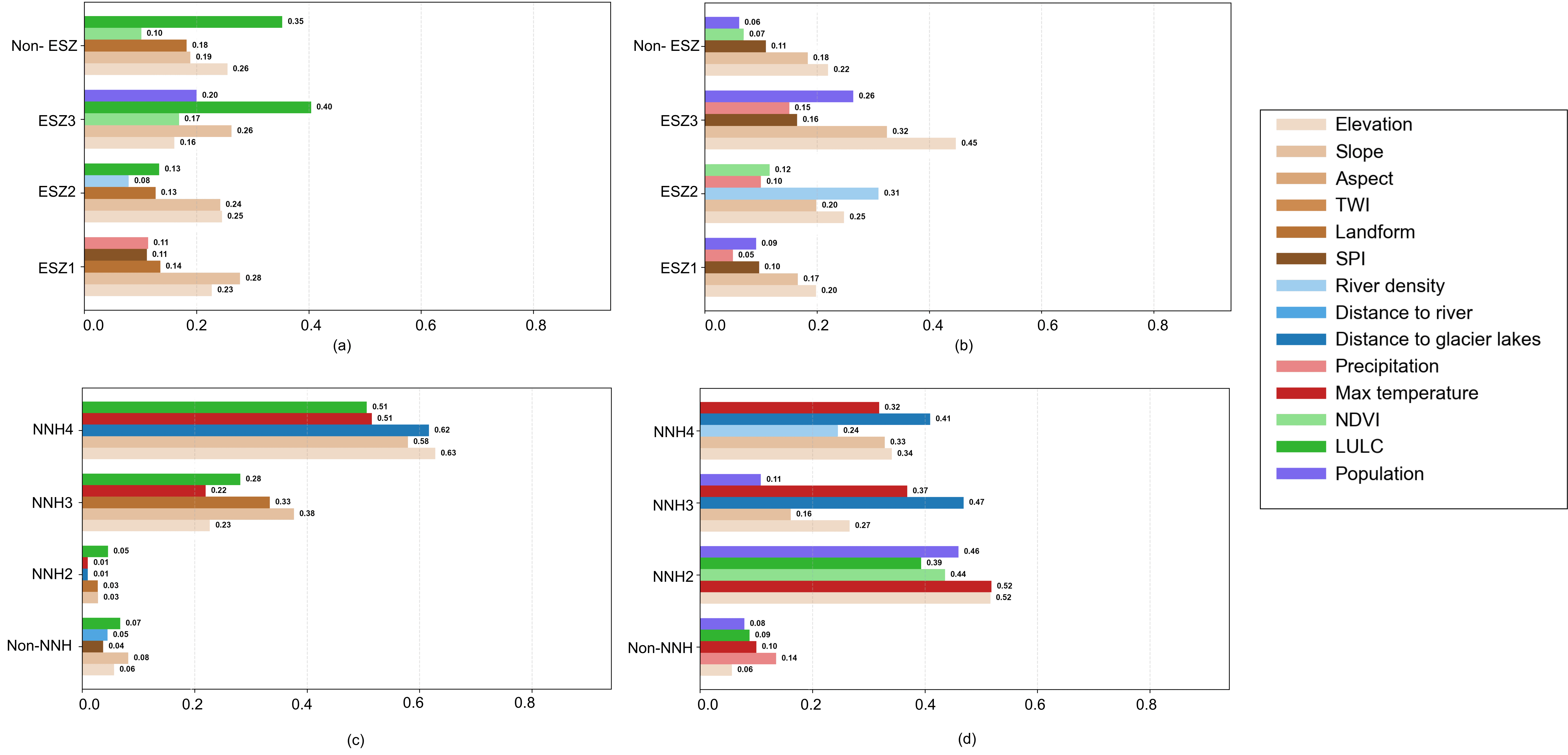}
    \caption{Factor Detector q-Statistic for Flood and Landslide Susceptibility for MLP-MVG,  (a) Kerala - Flood,  (b) Kerala - Landslide, (c) Nepal - Flood, (d) Nepal - Landslide}
    \label{fig:factordetector}
\end{figure*}

\begin{figure*}[!t]
    \centering
     \includegraphics[width=0.8\textwidth]{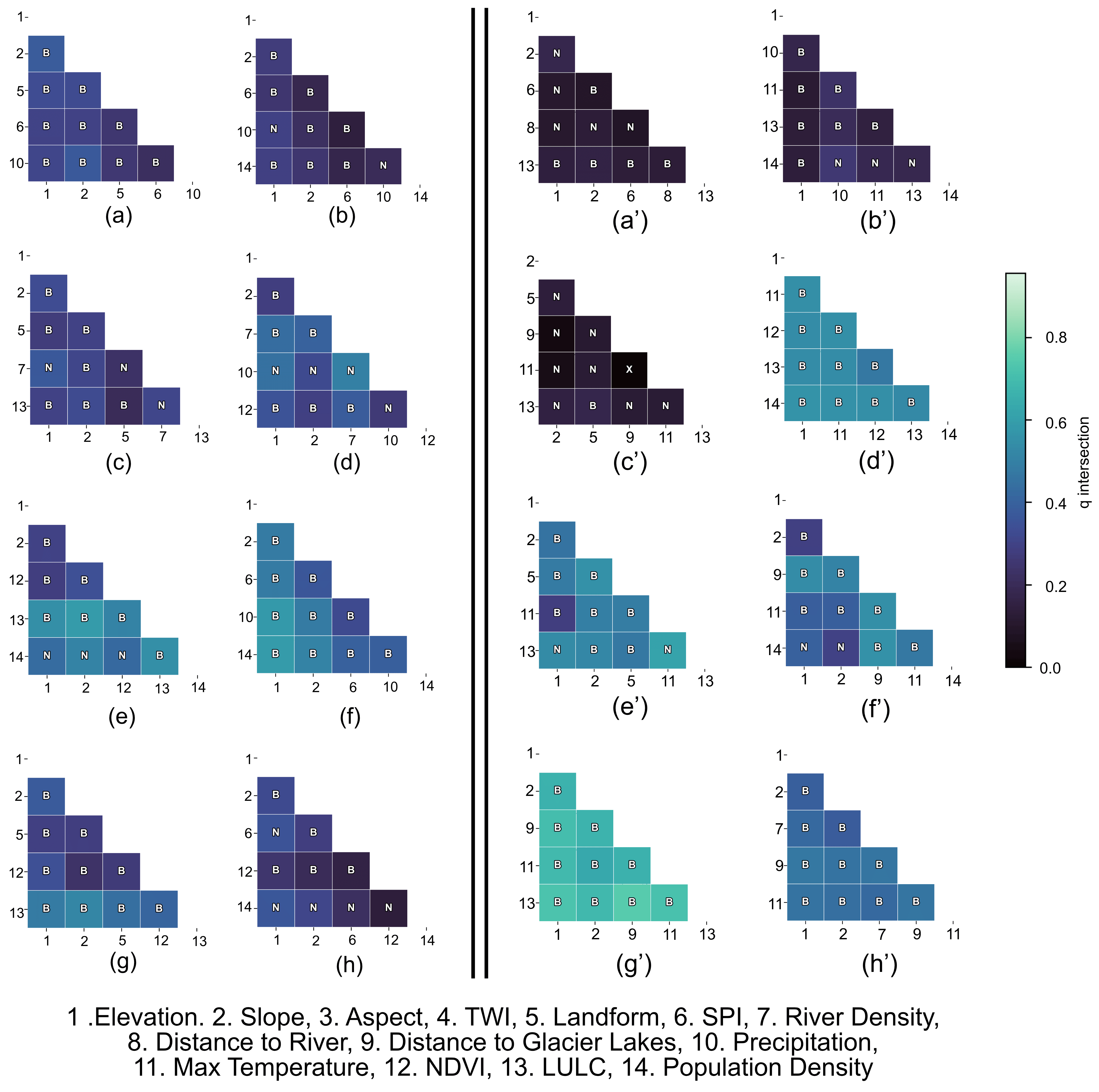}
    \caption{Top-5 factor interaction q-statistics obtained using the MLP-MVG model across contextual zones in Kerala and Nepal. For Kerala, panels (a), (c), (e), and (g) correspond to flood susceptibility for ESZ1, ESZ2, ESZ3, and Non-ESZ, respectively, whereas panels (b), (d), (f), and (h) show the corresponding landslide susceptibility. Similarly, for Nepal, panels (a$'$), (c$'$), (e$'$), and (g$'$) correspond to flood susceptibility for Non-NNH, NNH2, NNH3, and NNH4, respectively, whereas panels (b$'$), (d$'$), (f$'$), and (h$'$) show the corresponding landslide susceptibility. The overlaid symbols indicate interaction types, where B denotes bi-factor enhancement or bi-enhancement, and N denotes nonlinear enhancement.}
    \label{fig:interaction}
\end{figure*}

\begin{figure*}[!t]
    \centering
     \includegraphics[width=0.8\textwidth]{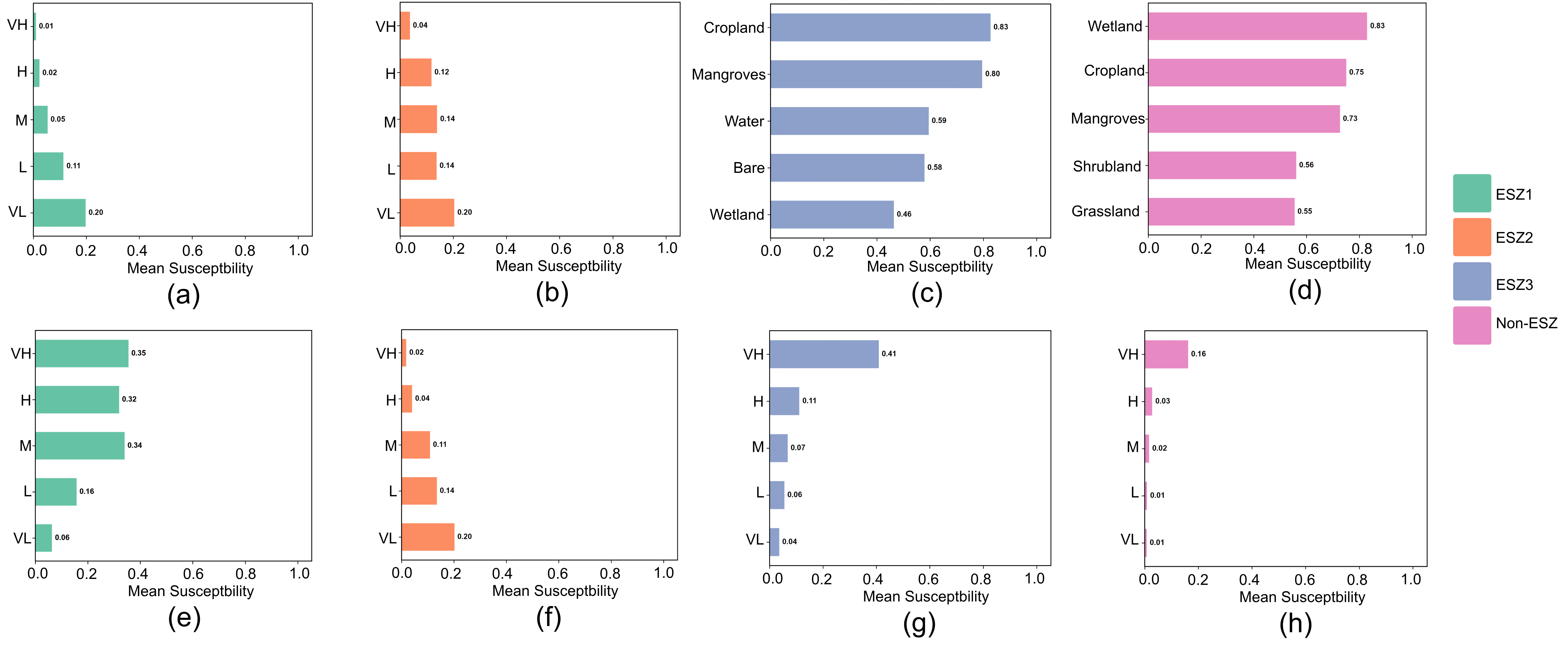}
    \caption{Risk detector results based on the highest q-statistic factor in each contextual zone of Kerala. Panels (a)-(d) show mean flood susceptibility for ESZ1, ESZ2, ESZ3, and Non-ESZ, respectively, whereas panels (e)-(f) show the corresponding mean landslide susceptibility.}
    \label{fig:riskdetector_kerala}
\end{figure*}

\begin{figure*}[!t]
    \centering
     \includegraphics[width=0.8\textwidth]{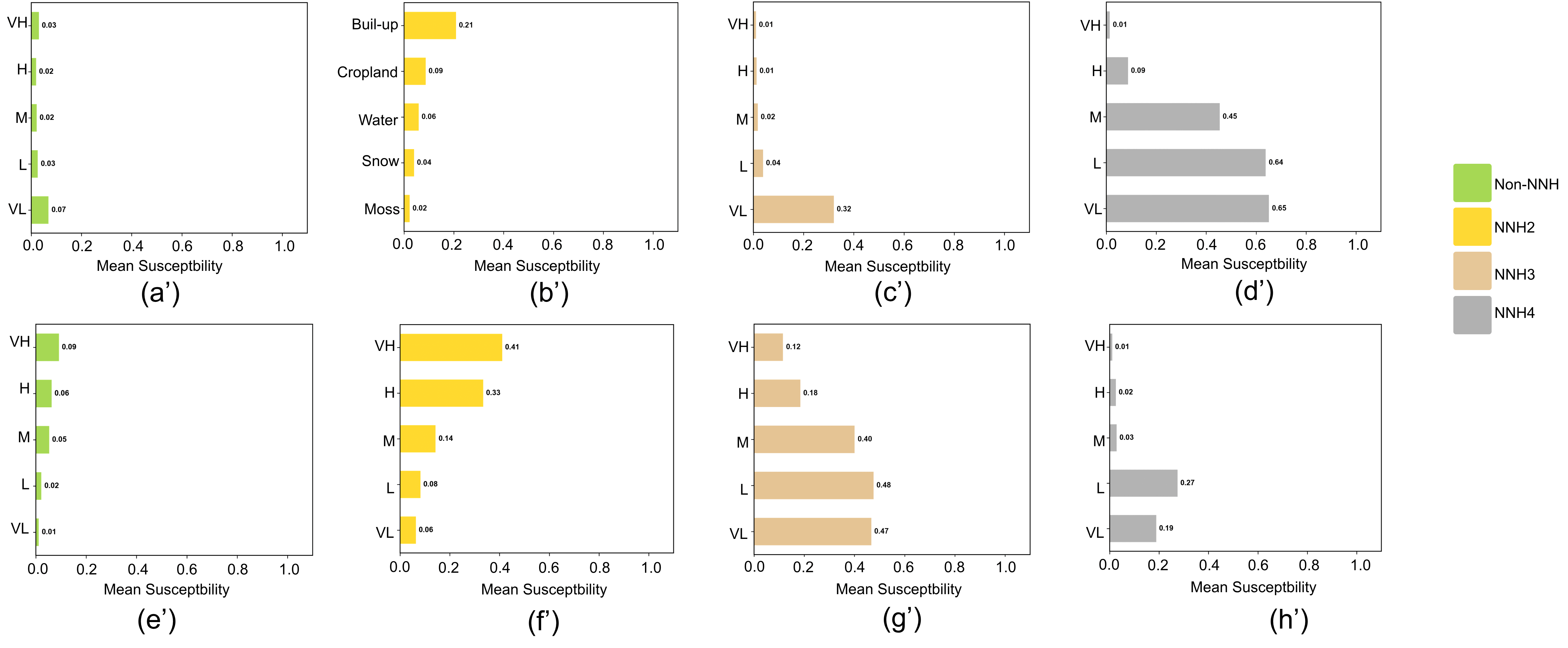}
    \caption{Risk detector results based on the highest q-statistic factor in each contextual zone of Nepal. Panels (a')-(d') show mean flood susceptibility for Non-NNH, NNH2, NNH3, and NNH4, respectively, whereas panels (e')-(h') show the corresponding mean landslide susceptibility.}
    \label{fig:riskdetector_nepal}
\end{figure*}

GeoDetector analysis was conducted to quantify spatial heterogeneity in the MoE outputs using a hierarchical screening framework. This analysis is done exclusively on MoE in FL-MHSM, as it showed the strongest overall predictive performance. This analysis identifies the dominant spatial drivers and interaction effects underlying the final susceptibility patterns. For each contextual zone, q-statistics were first computed for all conditioning variables. The five highest-ranked zone-specific factors were then selected for interaction analysis, while the top-ranked factor was further examined using the risk detector to evaluate stratified heterogeneity in susceptibility mapping. The top-ranked factor is further discretized to five classes, based on the range of values as \{VL,L,M,H,VH\}, similar to the severity levels of hazard susceptibility classes. Figure~\ref{fig:factordetector} gives the factor detector results, Figure~\ref{fig:interaction} gives the interaction results, and Figures~\ref{fig:riskdetector_kerala} and~\ref{fig:riskdetector_nepal} give the risk detector results for Kerala and Nepal, respectively.

\subsubsection{Kerala}
\paragraph{Dominant factors and interaction effects}
For flood susceptibility (Figure~\ref{fig:factordetector}), ESZ1 and ESZ2 were dominated by terrain-related controls, whereas ESZ3 and Non-ESZ were dominated by LULC. In ESZ1, slope ranked highest, followed by elevation, landform, precipitation, and SPI. The interaction structure in this zone (Figure~\ref{fig:interaction}) was entirely bi-enhanced, with the strongest joint effects concentrated in slope- and precipitation-related pairs, particularly slope \(\times\) precipitation and elevation \(\times\) precipitation.

In ESZ2, elevation was the dominant factor, followed by slope, landform, LULC, and river density. Most pairs in this zone were bi-enhanced, but the stronger interactions were concentrated in river-density- and LULC-related combinations, \eg~river density \(\times\) elevation, LULC \(\times\) elevation, and LULC \(\times\) slope. Nonlinear enhancement was observed for river density \(\times\) elevation, river density \(\times\) landform, and river density \(\times\) LULC.

By contrast, both ESZ3 and Non-ESZ were led by LULC. In ESZ3, LULC was followed by slope, population density, NDVI, and elevation. The strongest interactions were concentrated in LULC-related pairs, \eg~LULC \(\times\) elevation, LULC \(\times\) slope, and LULC \(\times\) NDVI, whereas nonlinear enhancement appeared mainly in population-related combinations involving population density with elevation, slope, and NDVI. Non-ESZ showed a similar LULC-led structure, followed by elevation, slope, landform, and NDVI. Here, all dominant-factor interactions were bi-enhanced, with the strongest joint effects again concentrated in LULC-mediated pairs.

For landslide susceptibility, elevation was the dominant factor in ESZ1, ESZ3, and Non-ESZ, whereas ESZ2 was distinguished by river density as the leading control. In ESZ1, elevation was followed by slope, SPI, population density, and precipitation. The interaction structure was mainly bi-enhanced, with stronger effects centered on elevation-related pairs, such as population density \(\times\) elevation and precipitation \(\times\) elevation. Nonlinear enhancement was limited to precipitation \(\times\) elevation and population density \(\times\) precipitation.

ESZ3 showed a similar elevation-led pattern, followed by slope, population density, SPI, and precipitation. All interactions among the dominant factors were bi-enhanced, with the strongest joint effects from elevation-related combinations, such as elevation \(\times\) population density, elevation \(\times\) precipitation, and elevation \(\times\) SPI.

Non-ESZ also remained elevation-dominated, followed by slope, SPI, NDVI, and population density. However, its interaction structure was more mixed, with stronger effects concentrated in elevation-coupled pairs and nonlinear enhancement appearing for SPI \(\times\) elevation and all population-related combinations.

ESZ2 contrasted with these zones, with river density ranking first, followed by elevation, slope, NDVI, and precipitation. This zone also showed a stronger nonlinear component. The larger interaction \(q\)-statistics were concentrated in river-density- and precipitation-related combinations, \eg~precipitation \(\times\) river density and river density \(\times\) elevation, while nonlinear enhancement appeared in precipitation \(\times\) elevation, precipitation \(\times\) slope, precipitation \(\times\) river density, and NDVI \(\times\) precipitation.

\paragraph{Risk detector}
The results (Figure~\ref{fig:riskdetector_kerala}) clarify how susceptibility varied across the strata of the top-ranked factor in each zone. For flood susceptibility, ESZ1 and ESZ2 showed the same directional pattern, with mean susceptibility decreasing in the five classes of the dominant factor, \ie~from VL to VH classes. In ESZ1, where slope was dominant, susceptibility was highest in the VL slope class and declined steadily toward the VH slope class, indicating that flood-prone areas were concentrated in gentler terrain. ESZ2 showed an analogous pattern for elevation, with the VL elevation class exhibiting the highest mean susceptibility and the VH class the lowest, consistent with greater flood susceptibility in lower-lying areas.

By contrast, the LULC-based results for ESZ3 and Non-ESZ reflected class-specific concentrations rather than ordinal gradients. In ESZ3, cropland and mangroves showed the highest mean susceptibility, followed by water, bare land, and wetland. In Non-ESZ, wetland ranked highest, followed by cropland and mangroves, whereas shrubland and grassland showed comparatively lower mean susceptibility.

For landslide susceptibility, ESZ1, ESZ3, and Non-ESZ all showed elevation-based risk structures in which higher elevation classes were associated with greater mean susceptibility. This trend was strongest in ESZ3, where the VH elevation class clearly dominated and the remaining classes showed a sharp decline. ESZ1 displayed a similar but less abrupt pattern, with the M and H classes also remaining significant. Non-ESZ followed the same general tendency, although more weakly, with the VH class again ranking highest and the lower classes showing only limited susceptibility.

ESZ2 differed from the other landslide zones because river density was the dominant factor. Here, susceptibility was highest in the VL river-density class, followed by the L and M classes, whereas the H and VH river-density classes showed much lower mean susceptibility.

\subsubsection{Nepal}
\paragraph{Dominant factors and interaction effects}
For flood susceptibility (Figure~\ref{fig:factordetector}), NNH4 and NNH3 were characterized by relatively strong topographic and cryospheric controls, whereas NNH2 and Non-NNH showed weaker or more mixed structures. In NNH4, elevation ranked highest, followed by distance to glacier lakes, slope, LULC, and maximum temperature. All interactions among these dominant factors were bi-enhanced (Figure~\ref{fig:interaction}), with the strongest joint effects concentrated in LULC-related pairs, particularly LULC \(\times\) elevation, LULC \(\times\) slope, LULC \(\times\) distance to glacier lakes, and LULC \(\times\) maximum temperature.

NNH3 was led by slope, followed by landform, LULC, elevation, and maximum temperature. Here again, the stronger interactions were concentrated in LULC-mediated combinations, especially LULC \(\times\) elevation, LULC \(\times\) slope, and LULC \(\times\) landform. Nonlinear enhancement was limited to LULC \(\times\) elevation and LULC \(\times\) maximum temperature.

By contrast, NNH2 showed the weakest individual explanatory strengths overall. LULC ranked first, followed by landform, slope, distance to glacier lakes, and maximum temperature. Its interaction structure was predominantly nonlinear and comparatively weak, with only LULC \(\times\) landform showing bi-enhancement, while most other pairs exhibited nonlinear enhancement or near-zero joint explanatory power.

Non-NNH showed slope as the leading factor, followed by LULC, elevation, distance to river, and SPI. The stronger joint effects were again concentrated in LULC-related pairs, particularly LULC \(\times\) elevation, LULC \(\times\) slope, and LULC \(\times\) SPI, whereas nonlinear enhancement appeared mainly in elevation- and distance-to-river-related combinations.

For landslide susceptibility, distance to glacier lakes was the dominant factor in both NNH3 and NNH4, while NNH2 and Non-NNH followed distinct patterns. In NNH4, distance to glacier lakes was followed by elevation, slope, maximum temperature, and river density. All interactions among the dominant factors were bi-enhanced, with stronger effects broadly distributed across river-density-, distance-to-glacier-lakes-, and maximum-temperature-related pairs.

NNH3 showed a similar dominant-factor structure, with distance to glacier lakes followed by maximum temperature, elevation, slope, and population. Its interaction matrix was likewise mainly bi-enhanced, with stronger effects centered on distance-to-glacier-lakes- and climate-related pairs, while nonlinear enhancement was limited to population \(\times\) elevation and population \(\times\) slope.

NNH2 differed from the other zones by showing co-dominance of elevation and maximum temperature, followed by population, NDVI, and LULC. All interactions in this zone were bi-enhanced, and the interaction \(q\)-statistics were generally high across the matrix, with the strongest joint effects concentrated in population-, NDVI-, and LULC-related combinations.

In Non-NNH, precipitation ranked highest, followed by maximum temperature, LULC, population, and elevation. The stronger joint effects were concentrated in elevation- and climate/LULC-related pairs, while nonlinear enhancement appeared for population \(\times\) precipitation, population \(\times\) maximum temperature, and population \(\times\) LULC.

\paragraph{Risk detector}
The results (Figure~\ref{fig:riskdetector_nepal}) show how susceptibility varied across the strata of the top-ranked factor in each Nepal zone. For flood susceptibility, Non-NNH and NNH3 both showed decreasing susceptibility from VL to VH slope classes, indicating concentration of flood-prone areas in gentler terrain. This trend was especially pronounced in NNH3, where the VL slope class was markedly higher than the others. NNH4 showed the same general pattern for elevation, with mean susceptibility highest in the VL and L elevation classes and declining sharply toward the high and VH classes, again indicating stronger flood susceptibility in lower-lying areas.

NNH2 contrasted with these ordinal patterns owing to LULC being its top factor. Susceptibility was concentrated in specific land-cover classes, with built-up areas showing the highest mean value, followed by cropland, water, snow, and moss.

For landslide susceptibility, NNH3 and NNH4 shared the same distance-to-glacier-lakes-based pattern, with susceptibility concentrated in the lower distance classes. In NNH3, the L and VL classes showed the highest mean susceptibility, followed by the M class, whereas the H and VH classes were much lower. NNH4 showed the same tendency, with the L class ranking highest and the VL class second.

Non-NNH followed a different pattern because precipitation was the dominant factor. Mean susceptibility increased toward the higher precipitation classes, reached its maximum in the VH class, and declined progressively toward the VL class. NNH2 was again distinct: the risk detector was applied to maximum temperature, one of the co-dominant top-ranked factors, and showed the highest mean susceptibility in the VH and H temperature classes, with progressively lower susceptibility in the M, L, and VL classes.

\section{Discussion} \label{sec:discussions}
Here, we discuss various structural consequences of fusion schemes that were applied in FL-MHSM, statistical analysis, and model interpretation.

\subsection{Comparative Analysis of Fusion and Ensemble Models}
The joint reading of the MHSM structure and the evaluation metrics shows that the fusion mechanism changes not only predictive discrimination, but also the way the final bivariate susceptibility pattern is organized. Across both Kerala and Nepal, EF, LF, and MoE differ in the balance they produce between hazard-separated regimes, intermediate compound regimes, and the concentration of extreme classes. This indicates that fusion is not merely a technical modeling choice, but a determinant of how flood and landslide relationships are expressed in the final FL-MHSM. This difference arises from whether the two hazards are learned within a shared representation, kept fully independent, or adaptively combined after separate learning.

EF contributes the most pronounced structural moderation. Its outputs reduce the dominance of strongly separated high- low combinations and redistribute susceptibility toward adjacent and intermediate compound classes, while the performance results show recurring gains in Recall and Brier score, especially for flood susceptibility. This indicates that EF adds value mainly by increasing continuity and moderation in the joint susceptibility space rather than by maximizing the sharpest hazard-wise separation. This pattern follows from the fact that joint learning places flood and landslide within the same probabilistic representation, thereby encouraging the model to organize susceptibility in a coupled space. Where the two hazards respond in a more aligned way to the same conditioning-factor space, this produces smoother and more continuous compound transitions; where their responses diverge more strongly, the same formulation suppresses hazard-specific gradients, which explains why EF does not uniformly retain the strongest single-hazard discrimination.

LF produces the opposite tendency. Its outputs preserve stronger high--low contrasts and sharper hazard-separated regimes, and this is accompanied by stronger or competitive hazard-wise discrimination, particularly in AUC-ROC and, in some cases, Brier score. This indicates that independent optimization is effective for preserving the most distinct single-hazard gradients, but that this strength is associated with a more polarized bivariate susceptibility structure. This pattern arises because LF learns flood and landslide independently and combines them only after prediction, so each hazard retains its own decision surface without being moderated by the other. As a result, LF preserves physiographic and hazard-specific contrasts more strongly, but it can also maintain sharper separation in the bivariate space than a more continuous compound organization would support.

MoE provides the strongest overall balance across predictive performance and structural stability. Structurally, it is closer to LF by preserving the main large-scale pattern of hazard separation, but it also reduces part of LF's stronger polarization. In performance terms, it is the strongest overall performer across regions and hazards rather than outperforming only in isolated cases. This indicates that spatially-adaptive fusion preserves much of the discriminative strength of LF while incorporating part of the structural moderation introduced by EF. This behavior follows from the fact that MoE does not impose a single shared representation or rely fully on independent prediction; instead, it combines both according to the local prediction context, retaining LF-like separation, where hazard-specific structure is strong, and EF-like moderation, where coupled behavior is more strongly expressed.

These patterns indicate that predictive performance and structural stability should be interpreted jointly. EF contributes value through moderation and improved sensitivity or calibration in several cases; LF is a strong baseline for hazard-wise discrimination, and MoE provides the most effective overall balance between these two behaviors. This indicates that multi-hazard susceptibility evaluation should not rely only on conventional accuracy-based metrics, because models with similar or stronger metric values can still produce materially different compound susceptibility structures. This requirement follows from the objective of FL-MHSM, which is not only to predict flood and landslide susceptibility individually, but also to generate a final bivariate map whose compound organization is interpretable and regionally consistent.

\subsection{Interpreting Spatial Heterogeneity in MoE Outputs}
The GeoDetector results show that the spatial organization of flood and landslide susceptibility varies across the contextual zones in both Kerala and Nepal. These differences indicate that susceptibility is shaped not only by the dominant factors within each zone, but also by how topography, land cover, drainage, and climate interact within different environmental settings.

\subsubsection{Kerala}
In Kerala, the flood-susceptibility pattern shows a clear reorganization of controls across the contextual zones. In ESZ1 and ESZ2, the dominant controls and risk-detector trends indicate that flooding is organized mainly by terrain. In ESZ1, the concentration of susceptibility in the very low slope class, together with the strong slope--precipitation and elevation--precipitation interactions, indicates that flood-prone areas are concentrated in gentler valley floors and local low-slope receiving zones where runoff from surrounding terrain converges under rainfall input. In ESZ2, the dominance of elevation and the highest susceptibility in the very low elevation class indicate a similar terrain-controlled pattern, with flooding organized around locally lower terrain positions and further shaped by drainage and land-cover effects. By contrast, ESZ3 and Non-ESZ show a different organization, with LULC emerging as the dominant control and the strongest interactions concentrated in LULC-related combinations. The concentration of higher susceptibility in cropland, mangroves, and wetlands indicates that flooding in these zones is controlled less by relief alone and more by land cover, surface wetness, and local drainage conditions. These patterns indicate that flood susceptibility in Kerala shifts from terrain-mediated runoff concentration in ESZ1 and ESZ2 to land-cover- and wet-environment-related organization in ESZ3 and Non-ESZ.

The landslide pattern in Kerala is more regionally consistent, but with a distinct zonal departure. In ESZ1, ESZ3, and Non-ESZ, elevation is the dominant factor, and susceptibility increases toward the higher elevation classes, indicating that landslide occurrence is concentrated mainly in elevated and relief-sensitive terrain. In ESZ1, the strong elevation-related interactions with precipitation and population density suggest that slope instability in these upland settings is further intensified by rainfall and local human pressure. ESZ3 shows the same elevation-led organization even more strongly, indicating that landslide-prone terrain is more tightly confined to upper relief settings where steep terrain and monsoon rainfall act together. Non-ESZ follows the same pattern more weakly, suggesting that landslide-prone areas there are restricted to scattered elevated and dissected pockets within a broader lower-relief background. ESZ2 departs from this structure. Here, river density becomes the dominant factor, and the strongest interactions involve precipitation and elevation, while the highest susceptibility occurs in the very low river-density class. This indicates that river density is not functioning as a simple monotonic control, but rather as an indicator of a drainage-conditioned slope setting in which rainfall and drainage configuration organize instability more strongly than elevation alone. From these, it indicates that slope instability is primarily concentrated in elevated terrain, with ESZ2 representing a more drainage-conditioned exception.

\subsubsection{Nepal}
In Nepal, the flood-susceptibility pattern is less uniform across the contextual zones and separates into more coherent and more fragmented settings. NNH3 and NNH4 show the clearest regional organization. In NNH4, the concentration of susceptibility in the low and very low elevation classes indicates that flooding is organized mainly in lower-lying terrain. In NNH3, susceptibility is highest in the very low slope class, indicating that flooding is concentrated in gentler valley and piedmont settings rather than on steep slopes themselves. In both zones, the strong interaction of LULC with elevation, slope, landform, or distance to glacier lakes indicates that topography provides the main setting for flood concentration, while land cover influences where water is retained or routed within that setting. By contrast, NNH2 shows a weaker and more localized flood structure. Its lower explanatory strength, predominantly nonlinear interactions, and concentration of susceptibility in specific land-cover classes, especially built-up areas, indicate that flood occurrence there is not organized by one strong regional terrain control, but by localized valley-scale and land-cover-dependent settings. Non-NNH also shows a weaker and more mixed pattern, where flood susceptibility remains concentrated in gentler terrain but without the same coherent regional structure seen in NNH3 and NNH4. Taken together, these patterns indicate that flood susceptibility in Nepal is most strongly organized in NNH3 and NNH4, whereas NNH2 and Non-NNH reflect more fragmented and locally conditioned flood settings.

Landslide susceptibility in Nepal shows a stronger overall organization than flood susceptibility, particularly in the higher contextual zones. In both NNH3 and NNH4, susceptibility is concentrated in the lower distance-to-glacier-lake classes, indicating that landslide-prone areas are organized around glacier-lake-related mountain catchments. The interaction structure supports this pattern, showing that landslide susceptibility in these zones is shaped jointly by glacier-lake proximity, elevation, slope, and climatic controls. This indicates that landslide occurrence in NNH3 and NNH4 is tied to steep upland terrain influenced by glacier-fed drainage and associated mountain hydrologic processes. NNH2 differs from these zones, but its landslide structure is not weak. The co-dominance of elevation and maximum temperature, together with strong bi-enhanced interactions involving population, NDVI, and LULC, indicates that landslide susceptibility there is organized by a combination of terrain, warmer local conditions, surface condition, and human-modified slopes rather than by glacier-lake-related controls alone. Non-NNH shows a different structure again, with precipitation as the dominant factor and the highest susceptibility in the very high precipitation class, indicating a more directly rainfall-driven landslide regime. Overall, the Nepal landslide pattern indicates that NNH3 and NNH4 retain the most coherent glacier- and mountain-related control, whereas NNH2 and Non-NNH reflect distinct but still physically meaningful landslide settings shaped by mountain-slope and rainfall-related processes, respectively.

\subsection{Interpreting Joint Uncertainty and Inter-Hazard Correlation}
Across both Kerala and Nepal, the predominantly negative inter-hazard correlation indicates that flood and landslide susceptibility do not increase together uniformly across space. This suggests that a strongly reinforcing joint hazard response is not the dominant background pattern in either study area. Instead, the EF model learns a joint susceptibility in which the two hazards often remain only weakly aligned, even when both are present within the same susceptibility regime.

The class-wise correlation shows that positive dependence is restricted rather than dominant, but its meaning differs between the two study areas. In Kerala, correlation remains negative across all represented flood-landslide classes, indicating that closer flood--landslide alignment does not emerge as a dominant class-level pattern even under the major compound regimes retained in the EF-MHSM. In Nepal, weak positive correlation appears only as a localized exception in the VL--M class, while the higher compound classes remain negatively correlated. This contrast shows that positive dependence and high compound susceptibility are not equivalent, and that the inter-hazard relationship is regime-dependent in both regions.

The uncertainty patterns support the same interpretation. In both study areas, uncertainty is generally lower in lower or more asymmetric susceptibility classes and higher in more balanced or stronger joint-susceptibility classes, although the pattern is class-dependent rather than strictly monotonic. Since uncertainty is represented by the joint term \(\log |\Sigma|\), these higher values should be interpreted as lower stability in the coupled estimate rather than as model failure. The uncertainty and correlation layers, therefore, extend the susceptibility map by showing not only where compound classes occur, but also where the EF estimate is more stable and where interpretation requires greater caution.

\subsection{Limitations and Future Directions} 
Susceptibility modeling was conducted using static conditioning variables and event inventories, without explicit representation of temporal cascade dynamics between hazards. Although correlation-aware modeling captures statistical interdependence at the output level, it does not explicitly simulate physical triggering mechanisms or sequential hazard processes. Incorporating time-resolved drivers or event-based cascade modeling could further refine multi-hazard risk representation.

The present FL-MHSM workflow focuses on two hazards. Extending the analysis to more number of hazards or different types of two-hazard systems will introduce additional complexity, requiring hazard-specific customization, scalable fusion strategies, and uncertainty-aware representations.

Multi-hazard class interpretation relies on the discretization of continuous susceptibility scores. While Jenks natural breaks provide data-driven classification, alternative discretization schemes may influence bivariate class proportions and spatial agreement measures. Future work could evaluate alternative sensitivity-based methods to determine classification thresholds. 

GeoDetector analysis was applied post hoc to interpret spatial drivers. Although this provides insight into conditioning mechanisms, driver coherence assessment was not directly integrated into model training or fusion selection. Future research could formalize landscape-driven fusion selection using quantitative driver heterogeneity indices or develop adaptive gating mechanisms informed by spatial partitioning.

\section{Conclusions} \label{sec:conclusion}
This study develops a spatially-adaptive FL-MHSM framework for regional flood--landslide susceptibility mapping by combining two-level spatial partitioning with Early Fusion, Late Fusion, and Mixture-of-Experts-based ensemble learning. The key conclusion is that the effect of fusion is region-dependent: different fusion strategies preserve different structural properties of the final multi-hazard pattern, and no single fusion mechanism can be assumed to be uniformly optimal across all heterogeneous settings. Early Fusion is useful for learning smoother and more continuous compound susceptibility organization, whereas Late Fusion better preserves sharper hazard-specific contrasts. The MoE ensemble results further show that these two strategies are not strictly competing alternatives, but complementary sources of information that can be adaptively combined to improve the final multi-hazard representation.

The broader contribution of this work is therefore not only a new workflow for a large-area/regional FL-MHSM, but also the demonstration that multi-hazard modeling should be interpreted as both a predictive task and a structural mapping problem. This perspective is particularly important for spatially heterogeneous landscapes, where the quality of a regional MHSM depends not only on metric performance, but also on how meaningfully the final bivariate susceptibility structure represents two-hazard compounding effects.

\bibliographystyle{unsrt}
\bibliography{papers_flsm}
\end{document}